\documentclass[12pt]{article}
\usepackage{amsmath,amsthm}
\usepackage{graphicx,subcaption}
\usepackage{enumerate}
\usepackage{natbib}
\usepackage{animate}
\usepackage{booktabs}
\usepackage{url} 
\usepackage[draft]{changes}

\RequirePackage[OT1]{fontenc}
\RequirePackage{amsthm,amsmath,bm,hyperref}
\newcommand{\pkg}[1]{{\normalfont\fontseries{b}\selectfont #1}}

\usepackage{undertilde}
\usepackage{graphicx}
\usepackage{float}
\usepackage{amsthm}
\usepackage{amsfonts}
\usepackage{amsmath,mathtools}
\usepackage{comment}
\DeclareMathAlphabet\mathbfcal{OMS}{cmsy}{b}{n} 
\usepackage{multirow}
\newtheorem{theorem}{Theorem}[section]
\newtheorem{corollary} {Corollary}[section]
\newtheorem{lemma} {Lemma}[section]
\newtheorem{proposition} {Proposition}[section]
\newcommand{\innp}[2]{\left\langle #1, #2 \right \rangle} 
\newcommand{\abs}[1]{\lvert #1 \rvert} 
\newcommand{\norm}[1]{\| #1 \|} 
\newcommand\numberthis{\addtocounter{equation}{1}\tag{\theequation}}

\usepackage{titlesec}

\titleformat{\paragraph}
{\normalfont\normalsize\bfseries}{\theparagraph}{1em}{}
\titlespacing*{\paragraph}
{0pt}{3.25ex plus 1ex minus .2ex}{1.5ex plus .2ex}

\begin{document}

\def\spacingset#1{\renewcommand{\baselinestretch}%
{#1}\small\normalsize} \spacingset{1}


\title{\bf Functional Time Series Forecasting: Functional Singular Spectrum Analysis Approaches}
\author{Jordan Trinka \\
Department of Mathematical and Statistical Sciences, \\Marquette University, USA\\
and \\ Hossein Haghbin\footnote{Contributed equally as much as first author}\\
Department of Statistics, \\ Persian Gulf University, Iran
\\
and \\
 Mehdi Maadooliat \\
Department of Mathematical and Statistical Sciences, \\Marquette University, USA\\}
\date{}
\maketitle
\vspace{-.1cm}

\begin{abstract}
In this paper, we propose two nonparametric methods used in the forecasting of functional time-dependent data, namely functional singular spectrum analysis recurrent forecasting and vector forecasting. Both algorithms utilize the results of functional singular spectrum analysis and past observations in order to predict future data points where recurrent forecasting predicts one function at a time and the vector forecasting makes predictions using functional vectors. We compare our forecasting methods to a gold standard algorithm used in the prediction of functional, time-dependent data by way of simulation and real data and we find our techniques do better for periodic stochastic processes.
\end{abstract}

\noindent%
\vfill

\newpage
\spacingset{1.5} 
\section{Introduction} \label{sec:int}
Functional data analysis (FDA) is a growing field of statistics that is showing promising results in analysis due to the fact that functional algorithms act on possibly more informative and smooth data. Often times statistical techniques that act on real-valued scalars or vectors are extended into the functional realm to handle such curved data. One example is principal component analysis (PCA) which was extended into functional PCA (FPCA) and multivariate FPCA so that dimension reduction may be performed on time-independent functional observations and many variants of these methods have been developed, see \cite{ramsay2005}, \cite{chiou2014}, and \cite{happ2018} for more details. Another example of this concept can be seen in singular spectrum analysis (SSA) \citep{golyandina2001} which is a decomposition technique for time series. The SSA algorithm was extended into functional SSA (FSSA) in \cite{haghbin2019}. They showed that the FSSA algorithm outperforms SSA and FPCA-based approaches in separating out sources of variation for smooth, time-dependent, functional data which is defined as a functional time series (FTS). In addition to SSA being extended to FSSA, the multivariate SSA (MSSA) approaches \citep{golyandina2015, hassani2013} have also been extended to the functional realm in \cite{trinka2020} where modeling done on a multivariate FTS of intraday temperature curves and images of vegetation in a joint analysis gave more prominent results. 

An important problem often confronted by researchers is prediction of stochastic processes. \cite{golyandina2001} expanded the results of the SSA and MSSA algorithms to deliver two commonly used nonparametric techniques in forecasting, called (a) SSA recurrent forecasting, and (b) SSA vector forecasting algorithms. Since the aforementioned SSA techniques have seen success in forecasting time series data, one may seek expanding that to the functional world for forecasting FTS.

One of the first approaches to FTS forecasting is given in \cite{hyndman2007} who found success in predicting mortality rates data. The method of \cite{hyndman2007} has been extended so that more recent FTS observations play a larger role in forecasts \citep{hyndman2009}. In addition, extensions have been made so that the method is robust in the presence of outliers \citep{shang2019,beyaztas2019}. The approach of \cite{hyndman2007} has also inspired a functional extension to the ARMAX model \cite{gonzalez2018} allowing for estimation of moving average terms.  In addition, the methodology of \cite{hyndman2007} and its variants have seen success in applications other than just mortality rate data, see \cite{shang2013} and \cite{wagner2018}. 

The approach of \cite{hyndman2007} and its extensions consists of two steps. In the first step, they use FPCA, or its variants, to reduce the dimensionality of the functional data and project the curves onto the FPCA basis. In the second step, they perform forecasting of the basis coefficient using various techniques such as ARIMA model. The details of the technique of \cite{hyndman2007} are given in the supplementary material. One may argue the time-dependency is not considered in the first step of their algorithm. In this paper we develop two forecasting algorithms based on FSSA that can incorporates the time-dependency into the decomposition of FTS. Furthermore, the proposed algorithms do not need the stationary assumption.



\begin{figure}[H]
\begin{center}
    \begin{subfigure}[b]{0.42\textwidth}
	\includegraphics[page=1,width=\textwidth]{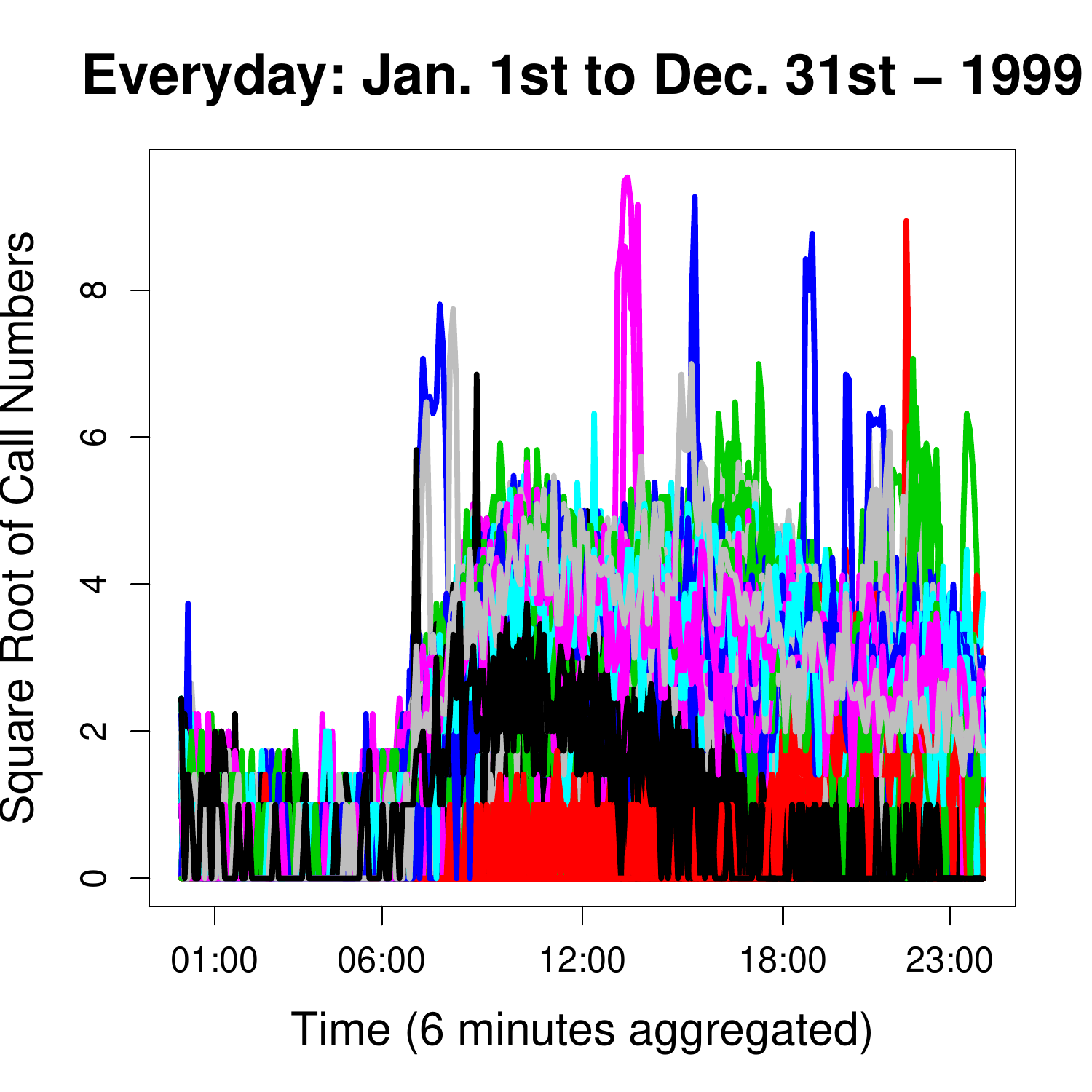}
    \end{subfigure}
    \begin{subfigure}[b]{0.42\textwidth}
	\includegraphics[page=2,width=\textwidth]{call_center_raw}
    \end{subfigure}
\caption{The square root of the number of calls to a call center}
\label{fig:motivating}
\end{center}
\end{figure}

In order to depict the idea of our approach and to show its utility, consider the following motivating example involving a real dataset which is described in detail in Subsection \ref{subsec:real}. Here we consider a FTS of $365$ curves where each function describes the square root of the number of calls, aggregated every six minutes, to a call center, in a day between January 1st, 1999 to December 31st, 1999 given in Figure \ref{fig:motivating}.

\begin{figure}[H]
\begin{center}
	\includegraphics[page=1,width=.38\textwidth]{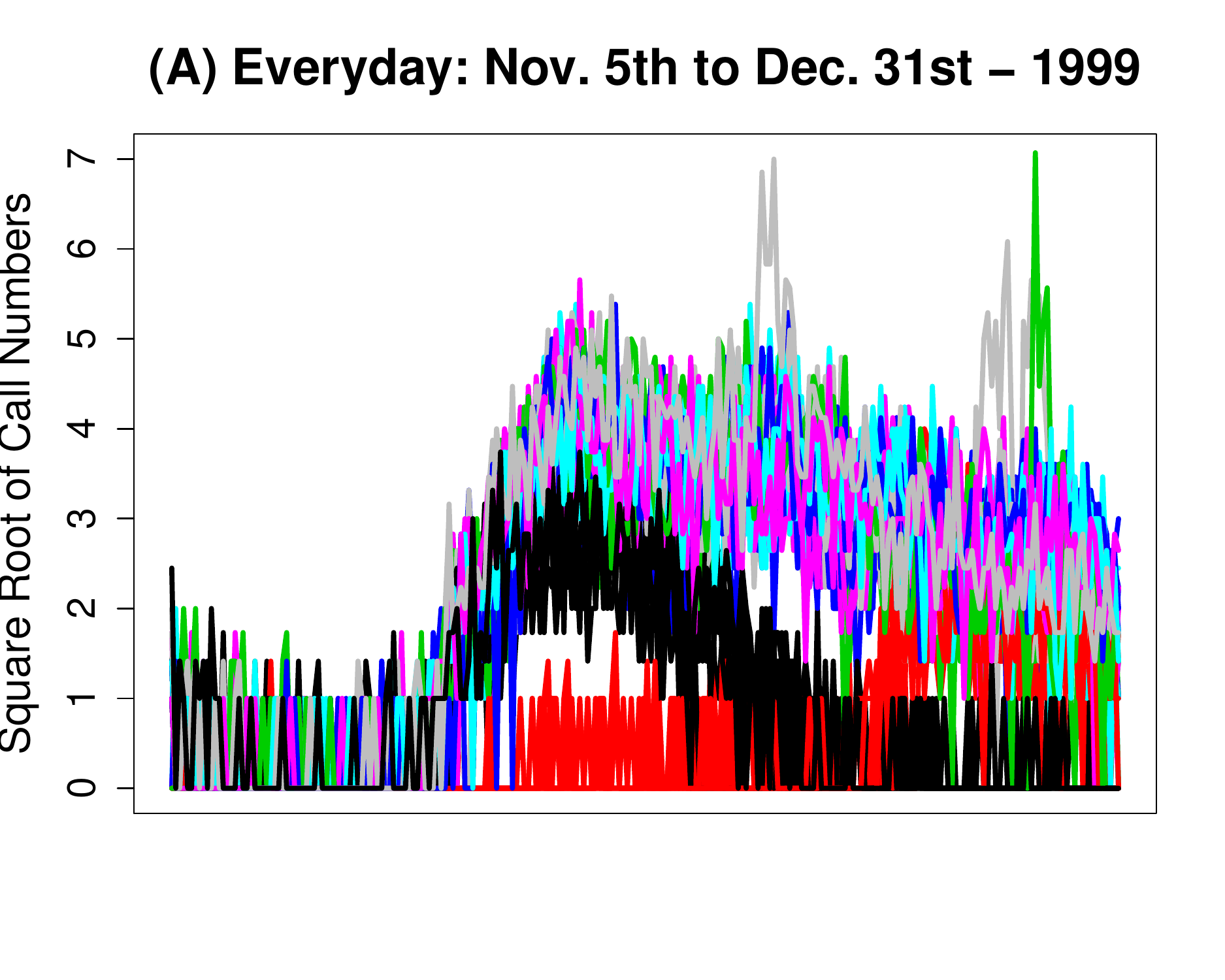}
		\vspace{-1cm}
		\hspace{-.5cm}
	\includegraphics[page=2,width=.38\textwidth]{call_center_just_predictions_5_signal}
	
	\includegraphics[page=3,width=.38\textwidth]{call_center_just_predictions_5_signal}
	\hspace{-.5cm}
	\includegraphics[page=4,width=.38\textwidth]{call_center_just_predictions_5_signal}
\caption{(A): Observed testing set; (B): Predictions using method of \cite{hyndman2007}; (C): FSSA vector forecasting predictions; (D): FSSA recurrent forecasting predictions.} 
\label{fig:motivating_call_2}
\end{center}
\end{figure}

\noindent Researchers have analyzed this data several times using FPCA-based approaches, \citep[see e.g.,][]{shen2005analysis,huang2008functional,maadooliat2015integrating}. The data also has been analyzed in the FSSA approach of \cite{haghbin2019}. Here we see that there exists a strong weekly periodicity in the data between workdays (Sunday through Thursday) and non-workdays (Friday and Saturday). We partition the $N=365$ curves into a training set of size $M=308$ (curves observed starting January 1, 1999 and ending November 4, 1999) and a testing set of the remaining $57$ (curves observed starting November 5, 1999 and ending December 31, 1999) functions. We compare our proposed methodologies of FSSA recurrent forecasting and FSSA vector forecasting to the approach of \cite{hyndman2007} in a rolling forecast fashion to obtain Figure \ref{fig:motivating_call_2}.

\noindent We see that the popular method of \cite{hyndman2007} struggles in predicting the periodic behavior of the call center data especially when trying to differentiate workdays from non-workdays, while our methods of FSSA recurrent and vector forecasting can capture this periodicity, and reflect that in the prediction.

The rest of the manuscript is organized as follows. In Section \ref{sec:theofssa}, we review the methodology of FSSA algorithm. In Section \ref{sec:forecast}, we develop the theoretical foundations of FSSA recurrent and vector forecasting. Then in Section \ref{sec:imp} we give the recipes needed to implement the recurrent and  vector forecasting approaches. Section \ref{sec:num} gives a simulation study and a real data study showing how our methods outperform a functional seasonal naive method and the popular technique of \cite{hyndman2007} in forecasting periodic FTS. Finally, we end with a discussion on results in Section \ref{sec:discussion}.


\section{Foundations of FSSA}\label{sec:theofssa}
We begin with some notations leveraged in the FSSA routine \citep{haghbin2019}. We consider $\mathbf{y}_N=\left(y_{1},\dots,y_{N}\right)$ as a FTS of length $N$ such that each $y_{i}:\left[0,1\right] \rightarrow \mathbb{R}$ belongs to $\mathbb{H}=\mathcal{L}^{2}\left(\left[0,1\right]\right)$ where $\mathcal{L}^{2}\left(\left[0,1\right]\right)$ is the space of square integrable real functions defined on the interval $\left[0,1\right]$. For some $x,y$ in the Hilbert space, $\mathbb{H}$, we have the that the inner product equipped to $\mathbb{H}$ is given by $\innp{x}{y}_{\mathbb{H}}=\int_{0}^{1}x\left(s\right)y\left(s\right)ds$. For a positive integer $m$, the space, $\mathbb{H}^{m}$, denotes the Cartesian product of $m$ copies of $\mathbb{H}$; such that any $\pmb{x} \in \mathbb{H}^{m}$ has the form $\pmb{x}\left(s\right)=\left(x_{1}\left(s_{1}\right),x_{2}\left(s_{2}\right),\dots,x_{m}\left(s_{m}\right)\right)^{\top}$, where $x_{i} \in \mathbb{H}$, $\mathbf{s}=\left(s_{1},s_{2},\dots,s_{m}\right)$ and $s_{i} \in \left[0,1\right]$. For some $\pmb{x}, \pmb{y}$ in the Hilbert space, $\mathbb{H}^{m}$, we define the inner product as $\innp{\pmb{x}}{\pmb{y}}_{\mathbb{H}^{m}}=\sum_{i=1}^{m}\innp{x_{i}}{y_{i}}_{\mathbb{H}}$. We denote the norms by $\norm{\cdot}_{\mathbb{H}}$ and $\norm{\cdot}_{\mathbb{H}^{m}}$ in the spaces $\mathbb{H}$ and $\mathbb{H}^{m}$ respectively. Given $x,y \in \mathbb{H}$, then the tensor (outer) product of these two elements gives us the operator $x \otimes y:\mathbb{H} \rightarrow \mathbb{H}$ where for some $h \in \mathbb{H}$, we have $\left(x \otimes y \right)h=\innp{x}{h}_{\mathbb{H}}y$.

For positive integers $L$ and $K$, we denote $\mathbb{H}^{L \times K}$ as the linear space spanned by operators $\mathbfcal{Z}:\mathbb{R}^{K} \rightarrow \mathbb{H}^{L}$, specified by $\left[z_{i,j}\right]_{i=1,\dots,L}^{j=1,\dots,K}$ where

\begin{equation*}
\mathbfcal{Z}\pmb{a}=\begin{pmatrix}\sum_{j=1}^{K}a_{j}z_{1,j} \\ \vdots \\ \sum_{j=1}^{K}a_{j}z_{L,j} \end{pmatrix}, \quad z_{i,j} \in \mathbb{H}, \text{ and }  \pmb{a}=\left(a_{1},\dots,a_{K}\right)\in \mathbb{R}^{K}.
\end{equation*}

\noindent We call an operator $\tilde{\mathbfcal{Z}} = \left[\tilde{z}_{i,j}\right] \in \mathbb{H}^{L \times K}$ Hankel if $\norm{\tilde{z}_{i,j}-g_{s}}_{\mathbb{H}}=0$ for some $g_{s} \in \mathbb{H}$, where $s=i+j$. The space of such Hankel operators will be denoted with $\mathbb{H}_{H}^{L \times K}$. For two given operators $\mathbfcal{Z}_{1}=\left[z_{i,j}^{\left(1\right)}\right]_{i=1,\dots,L}^{j=1,\dots,K}$ and $\mathbfcal{Z}_{2}=\left[z_{i,j}^{\left(2\right)}\right]_{i=1,\dots,L}^{j=1,\dots,K}$ in $\mathbb{H}^{L \times K}$, define

\begin{equation*}
\innp{\mathbfcal{Z}_{1}}{\mathbfcal{Z}_{2}}_{\mathbfcal{F}}=\sum_{i=1}^{L}\sum_{j=1}^{K}\innp{z_{i,j}^{\left(1\right)}}{z_{i,j}^{\left(2\right)}}_{\mathbb{H}}.
\end{equation*}

\noindent It follows immediately that $\innp{\cdot}{\cdot}_{\mathbfcal{F}}$, defines an inner product on $\mathbb{H}^{L \times K}$. We will call it Frobenius inner product of two operators in $\mathbb{H}^{L \times K}$. The associated Frobenius norm is $\norm{\mathbfcal{Z}}_{\mathbfcal{F}}=\sqrt{\innp{\mathbfcal{Z}}{\mathbfcal{Z}}_{\mathbfcal{F}}}$. Now we jump into the FSSA procedure with the goal of extracting time-dependent modes of variation that can be used for forecasting.

\subsection{Review of the FSSA Algorithm}

For an integer $1 < L < N/2$, let $K=N-L+1$ and define a set of multivariate functional vectors in $\mathbb{H}^{L}$ by

\begin{equation*}
\pmb{x}_{j}=\left(y_{j},y_{j+1},\dots,y_{j+L-1}\right),\quad j=1,\dots,K,
\end{equation*}

\noindent where $\pmb{x}_{j}$'s denote the functional $L$-lagged vectors. We also have that $L$ should be chosen to be a multiple of the periodicity present in the data \citep{haghbin2019}. We now continue with the FSSA methodology.

\subsubsection*{Step 1. Embedding}
In this first step we form the trajectory operator, $\mathbfcal{X}:\mathbb{R}^{K} \rightarrow \mathbb{H}^{L}$, which is defined by

\begin{equation}\label{trajop}
\mathbfcal{X}\pmb{a}=\sum_{j=1}^{K}a_{j}\pmb{x}_{j}, \quad \pmb{a} \in \mathbb{R}^{K}.
\end{equation}

\noindent We see that the rank of $\mathbfcal{X}$ is $0<r\leq K$ and that the range of this operator includes all possible linear combinations of the functional $L$-lagged vectors. This step of embedding can also be viewed as applying the invertible operation $\mathbfcal{T}:\mathbb{H}^{N} \rightarrow \mathbb{H}_{H}^{L \times K}$ such that $\mathbfcal{T}\left(\mathbf{y}_{N}\right)=\mathbfcal{X}$.

\subsubsection*{Step 2. Decomposition}
Since $\mathbfcal{X}$ is of finite rank, then there exists orthonormal elements $\{\pmb{\psi}_{i}\}_{i=1}^{r}$ from $\mathbb{H}^{L}$ and orthonormal vectors $\{\pmb{v}_{i}\}_{i=1}^{r}$ from $\mathbb{R}^{K}$ such that

\begin{equation*}
\mathbfcal{X}\pmb{a}=\sum_{i=1}^{r}\sqrt{\lambda_{i}}\innp{\pmb{v}_{i}}{\pmb{a}}_{\mathbb{R}^{K}}\pmb{\psi}_{i}, \quad \text{for all } \pmb{a} \in\mathbb{R}^{K}, \nonumber
\end{equation*}

\noindent where $\lambda_{i}$'s are non-ascending positive scalars. We refer to the previous equation as the functional singular value decomposition (fSVD) of $\mathbfcal{X}$. The $\sqrt{\lambda_{i}}$ is the $i^{\text{th}}$ singular value, $\pmb{\psi}_{i}$ is the $i^{\text{th}}$ left singular function, and $\pmb{v}_{i}$ is the $i^{\text{th}}$ right singular vector of the trajectory operator. Therefore we call the collection $\{\sqrt{\lambda_{i}},\pmb{\psi}_{i},\pmb{v}_{i}\}$ as the $i^{\text{th}}$ eigentriple of $\mathbfcal{X}$. One can write $\mathbfcal{X}$ as

\begin{equation*}
\mathbfcal{X}=\sum_{i=1}^{r}\sqrt{\lambda_{i}}\pmb{v}_{i}\otimes \pmb{\psi}_{i}. \nonumber
\end{equation*}

\subsubsection*{Step 3. Grouping}

We refer the readers to \cite{haghbin2019} for the general discussion in the grouping step. In this work, we combine rank one operators resulted from fSVD of $\mathbfcal{X}$ together for reconstructing the signal. Consider $\mathfrak{S}$ to be a subset of indices, $\{1,\dots,r\}$, and define $\mathbfcal{X}_\mathfrak{s}=\sum_{i\in\mathfrak{S}}\sqrt{\lambda_{i}}\pmb{v}_{i}\otimes \pmb{\psi}_{i}$. This allows us to write
\begin{equation*}
\mathbfcal{X} = \mathbfcal{X}_{\mathfrak{s}}+\mathbfcal{X}_{\mathfrak{n}},\nonumber
\end{equation*}
where $\mathbfcal{X}_{\mathfrak{n}}=\sum_{i\in\mathfrak{S}^c}\sqrt{\lambda_{i}}\pmb{v}_{i}\otimes \pmb{\psi}_{i}$ is the residual component, and $\mathbfcal{X}_{\mathfrak{s}}$ represents the signal component.

\subsubsection*{Step 4. Reconstruction}

In this step, we would like to use $\mathbfcal{T}^{-1}:\mathbb{H}_{H}^{L \times K} \rightarrow \mathbb{H}^{N}$ to transform back $\mathbfcal{X}_\mathfrak{s}$ to a FTS, $\tilde{\mathbf{y}}_{N} \in \mathbb{H}^{N}$, that captures the deterministic nature of $\mathbf{y}_N$. To do this, we first perform an orthogonal projection of $\mathbfcal{X}_\mathfrak{s}$ onto $\mathbb{H}_{H}^{L \times K}$, which is a closed subspace of $\mathbb{H}^{L \times K}$, with respect to the Frobenius norm. We denote the elements of $\mathbfcal{X}_\mathfrak{s}$ and the corresponding orthogonal projection, $\tilde{\mathbfcal{X}} \in \mathbb{H}_{H}^{L \times K}$, with $\left[x_{i,j}^\mathfrak{s}\right]$ and $\left[\tilde{x}_{i,j}\right]$, respectively. Then, using the results of \cite{haghbin2019} we have that

\begin{equation*}
\tilde{x}_{i,j}=\frac{1}{n_{s}}\sum_{\left(k,l\right):k+l=s}x_{k,l}^\mathfrak{s}, \nonumber
\end{equation*}

\noindent where $s=i+j$ and $n_{s}$ stands for the number of $\left(l,k\right)$ pairs such that $l+k=s$. We denote this projection by $\pmb{\Pi}_{H}:\mathbb{H}^{L \times K} \rightarrow \mathbb{H}_{H}^{L \times K}$ such that 

\begin{equation}\label{eqn:xs2xt}
\tilde{\mathbfcal{X}}=\pmb{\Pi}_{H}\mathbfcal{X}_\mathfrak{s}. 
\end{equation}

Similar to \eqref{trajop}, the operators $\tilde{\mathbfcal{X}}$ and $\mathbfcal{X}_\mathfrak{s}$ can be  written as linear combinations of  functional $L$-lagged vectors. We denote those lag-vectors as $\tilde{\pmb{x}}_{j}$ and $\pmb{x}^\mathfrak{s}_{j}$ respectively. From \eqref{eqn:xs2xt}, we find that $\tilde{\mathbf{y}}_{N}=\mathbfcal{T}^{-1}\tilde{\mathbfcal{X}}=\left(\tilde{y}_{1},\dots,\tilde{y}_{N}\right)$. 

As an immediate result, we have that the residual elements of the signal, $\mathbf{y}_{N}$, are captured in $\mathbf{y}_{N}-\tilde{\mathbf{y}}_{N}$ and we typically do not include these noisy modes of variation in the SSA-based forecasts, see \cite{golyandina2001, golyandina2013, golyandina2015, hassani2013} for more information.


\section{FSSA Forecasting}\label{sec:forecast}
The following contains some notation that will be leveraged throughout the rest of this Section and Section \ref{sec:imp}. For each functional vector $\pmb{x} \in \mathbb{H}^{L}$, denote by $\pmb{x}^{\nabla} \in \mathbb{H}^{L-1}$ and $\pmb{x}^{\Delta} \in \mathbb{H}^{L-1}$ as the functional vectors consisting of the first and the last (respectively) $L-1$ components of the vector $\pmb{x}$. Setting $k<r$, let us define $\mathbb{L}=\text{sp}\{\pmb{\psi}_{i}\}_{i=1}^{k}$ and $\mathbb{L}^{\nabla}=\text{sp}\{\pmb{\psi}_{i}^{\nabla}\}_{i=1}^{k}$. Moreover, let $\pi_{i} \in \mathbb{H}$ be the last component of the functional vector $\pmb{\psi}_{i}$. With this notation, we may continue into the theory of forecasting.

Here, we offer an important theorem and a corollary that will be leveraged in the FSSA recurrent and vector forecasting algorithms. All proofs of presented theory can be found in the supplementary materials.

\begin{theorem}\label{thm:nunorm}
Let $\mathbb{E}:=\{\left(\underline{0},\dots,\underline{0},x\right)\in \mathbb{H}^{L}|x \in \mathbb{H}\}$ and define the operator $\mathbfcal{V}:\mathbb{H} \rightarrow \mathbb{H}$ such that $\mathbfcal{V}:=\sum_{i=1}^{k} \pi_{i} \otimes \pi_{i}$. If $\mathbb{E} \cap \mathbb{L}=\emptyset$, then we have that

\begin{equation}
\norm{\mathbfcal{V}} = \underset{\norm{x}_{\mathbb{H}}=1}{\sup}\norm{\mathbfcal{V}\left(x\right)}_{\mathbb{H}} < 1,  \nonumber
\end{equation}

\noindent where $\norm{\cdot}$ is the operator norm.

\end{theorem}

\begin{corollary}\label{cor:inv}
Given that the conditions of Theorem \ref{thm:nunorm} are met, we have that $\sum_{l=0}^{\infty} \mathbfcal{V}^{l}$ is a convergent Neumann series and that $\left(\mathbfcal{I}-\mathbfcal{V}\right)^{-1}$ exists. Furthermore, we obtain the equality
\begin{equation}
\left(\mathbfcal{I}-\mathbfcal{V}\right)^{-1}=\sum_{l=0}^{\infty} \mathbfcal{V}^{l}, \nonumber
\end{equation}
where $\mathbfcal{I}:\mathbb{H}\rightarrow \mathbb{H}$ is the identity operator.
\end{corollary}

\noindent The primary assumption seen in Theorem \ref{thm:nunorm} is that for any $\pmb{z} \in \mathbb{E}$, we must have $\pmb{z} \not \in \mathbb{L}$. Now we define $\mathbf{g}_{N+M}=\left(g_{1},\dots,g_{N},g_{N+1},\dots,g_{N+M}\right) \in \mathbb{H}^{N+M}$ as a FTS of length $N+M$, where in the following algorithms (recurrent and vector forecasting), the first $N$ elements, $\{g_i\}_{i=1}^{N}$, are close to $\tilde{\mathbf{y}}_N$, and the main goal is to predict the last $M$ terms ($g_{N+1}$ to $g_M$).

\subsection{FSSA Recurrent Forecasting Algorithm}
The SSA recurrent forecasting approach of \cite{golyandina2001} leverages the fact that the next forecasted value of a reconstructed time series can be expressed as a linear combination of the previous $L-1$ elements. When we migrate to the functional realm, we move towards linear combinations of functions where the coefficient scalar weights are replaced with operators that perform the weighting. With this in mind, we now give the following theorem which we will employ in the FSSA recurrent and vector forecasting algorithms. 

\begin{theorem}\label{thm:rforecast}
Given that the conditions of Theorem \ref{thm:nunorm} are met, for any $\pmb{y}=\left(y_{1},\dots,y_{L}\right)\in\mathbb{L}$, the last component, $y_{L}$, can be expressed by a linear combination of the previous $L-1$ components:

\begin{equation}
y_{L}=\sum_{j=1}^{L-1}\mathbfcal{A}_{j}y_{j},\nonumber
\end{equation}

\noindent where $\mathbfcal{A}_{j}:\mathbb{H} \rightarrow \mathbb{H}$ is an operator given by $\mathbfcal{A}_{j}=\sum_{n=1}^{k}\psi_{j,n}\otimes \left(\mathbfcal{I}-\mathbfcal{V}\right)^{-1}\pi_{n}$, and $\psi_{j,n} \in \mathbb{H}$ is the $j^{\text{th}}$ component of the $n^{\text{th}}$ left singular function.

\end{theorem}


One may use the result of Theorem \ref{thm:rforecast}, and the background of the recurrent SSA algorithm given in \cite{golyandina2001} to present the following algorithm. \newline

\noindent \textit{Algorithm A - FSSA Recurrent Forecasting:}
\[g_{i}= \begin{cases} 
      \tilde{y}_{i} & i=1,\dots,N \\
     \sum_{j=1}^{L-1}\mathbfcal{A}_{j}g_{i+j-L} & i=N+1,\dots,N+M
   \end{cases}.\label{eqn:rforecast}\numberthis
\]
For brevity, we call this algorithm \textit{FSSA R-forecasting algorithm}.


\subsection{FSSA Vector Forecasting Algorithm}
The R-forecasting can be extended into an algorithm known as vector forecasting that allows us to perform prediction using functional $L$-lagged vectors. We first define the operator, $\mathbfcal{P}^{\nabla}:\mathbb{R}^{k} \rightarrow \mathbb{H}^{L-1}$ such that for any $\pmb{a} \in \mathbb{R}^{k}$, we have

\begin{equation*}
\mathbfcal{P}^{\nabla}\left(\pmb{a}\right)=\sum_{n=1}^{k}a_{n}\pmb{\psi}_{n}^{\nabla}.
\end{equation*}

\noindent It is easy to see the adjoint operator can be obtained as $\left(\mathbfcal{P}^{\nabla}\right)^{*}:\mathbb{H}^{L-1} \rightarrow \mathbb{R}^{k}$, such that for all $\pmb{v} \in \mathbb{H}^{L-1}$, we have

\begin{equation*}
\left(\mathbfcal{P}^{\nabla}\right)^{*}\left(\pmb{v}\right)=\begin{bmatrix}\innp{\pmb{v}}{\pmb{\psi}_{1}^{\nabla}}_{\mathbb{H}^{L-1}} \\ \vdots \\ \innp{\pmb{v}}{\pmb{\psi}_{k}^{\nabla}}_{\mathbb{H}^{L-1}} \end{bmatrix}. 
\end{equation*}

For a given vector $\pmb{x}\in\mathbb{L}$, the goal of vector forecasting is to find a linear operator $\mathbfcal{Q}:\mathbb{L}\rightarrow\mathbb{L}$, such that for some $\pmb{y}=\mathbfcal{Q}\pmb{x}$, the distance between the vectors $\pmb{x}^\Delta$ and $\pmb{y}^\nabla$ is minimal. The following Proposition can be used to obtain such a $\pmb{y}^\nabla$.

\begin{proposition}\label{prop:vforecastpi}
Given that the conditions of Theorem \ref{thm:nunorm} are met, the operator $\mathbf{\Pi}$ defined as
\begin{equation*}
\pmb{\Pi}:=\mathbfcal{P}^{\nabla}\Bigl(\left(\mathbfcal{P}^{\nabla}\right)^{*}\mathbfcal{P}^{\nabla}\Bigr)^{-1}\left(\mathbfcal{P}^{\nabla}\right)^{*}
\end{equation*}

\noindent is an orthogonal projection from $\mathbb{H}^{L-1}$ onto $\mathbb{L}^{\nabla}$.

\end{proposition}

\noindent As an immediate result of Proposition \ref{prop:vforecastpi}, we have $\pmb{y}^\nabla=\pmb{\Pi}\pmb{x}^\Delta$. Now, using Theorem \ref{thm:rforecast}, one may we obtain the last component of $\pmb{y}$ (i.e., $y_L=\sum_{j=1}^{L-1}\mathbfcal{A}_{j}y^\nabla_j$).
The following Proposition further simplify this expression.

\begin{proposition}\label{prop:vforecastlast}
\noindent Under the conditions of Theorem \ref{thm:nunorm}, we have

 \begin{equation*}
y_L= \sum_{j=1}^{L-1}\mathbfcal{A}_{j}{x}^{\Delta}_{j}.
\end{equation*}

\end{proposition}

\noindent Finally, one may derive the linear operator $\mathbfcal{Q}$ as following:

\begin{equation}\label{eqn:Q}
\mathbfcal{Q}\left(\pmb{x}\right):=\begin{pmatrix} \pmb{\Pi}\left(\pmb{x}^{\Delta}\right) \\ \sum_{j=1}^{L-1}\mathbfcal{A}_{j}x^{\Delta}_{j}\end{pmatrix}, \qquad {\pmb{x}} \in \mathbb{L}.
\end{equation}

\noindent\textit{Algorithm B - FSSA Vector Forecasting:}

\begin{enumerate}
\item[1.] Define the functional $L$-lagged vectors

\[\pmb{w}_{j}= \begin{cases} 
      \pmb{x}^{\mathfrak{s}}_{j} & j=1,\dots, K \\
     \mathbfcal{Q}\pmb{w}_{j-1} & j=K+1, \dots, K+M
   \end{cases}\label{eqn:vforecast}\numberthis
\]

\item[2.] Form the operator $\mathbfcal{W} \in \mathbb{H}^{L \times \left(K+M\right)}$ whose range is linearly spanned by the set $\{\pmb{w}_{i}\}_{i=1}^{K+M}$.

\item[3.] Hankelize $\mathbfcal{W}$ in order to extract the FTS $\mathbf{g}_{N+M}$.

\item[4.] The functions $g_{N+1}, \cdots, g_{N+M}$ form the $M$ terms of the FSSA vector forecast.

\end{enumerate}





\noindent  From here, we may refer to this algorithm as \textit{FSSA V-forecasting algorithm}. In the next Section, we present computer implementation of both algorithms.

\section{Implementation Strategy}\label{sec:imp}
In practice, functional data are being recorded discretely and then converted to functional objects using proper smoothing techniques. We refer to \cite{ramsay2005} for more details on preprocessing the raw data. Here, we start with some notation that will be leveraged in the implementation of the R-forecasting and V-forecasting algorithms. We assume that $\{\nu_{i}\}_{i=1}^{d}$ is a linearly independent basis for $\mathbb{H}_{d}$, which is a $d$-dimensional subspace of $\mathbb{H}$. For any $f \in \mathbb{H}_{d}$ there exists a unique vector, $\mathbf{c}_{f}=(c_{f,1},\cdots,c_{f,d})^\top \in \mathbb{R}^{d}$, where
$f=\sum_{i=1}^{d}{c}_{f,i}\nu_{i}. \nonumber$
\noindent We refer to $\mathbf{c}_{f}$ as the corresponding coefficients of $f$. Let $\mathbf{G}=\left[\innp{\nu_{i}}{\nu_{j}}_{\mathbb{H}}\right]_{i,j=1}^{d}$ to be the $d \times d$ Gram matrix. 

From hereafter, we consider all discussed operators whose domain or range are infinite dimensional Hilbert spaces to operate on or map to the corresponding $d$-dimensional subspace. Since the FSSA R-forecasting and V-forecasting algorithms are both dependent on $\left(\mathbfcal{I}-\mathbfcal{V}\right)^{-1}$, we explore one particular implementation of this operator in the following lemma that will be leveraged in the recipes for R-forecasting and V-forecasting.





\begin{lemma}\label{lem:opimp}
Define $\mathbf{D}=\mathbf{G}^{\frac{1}{2}}\left[\mathbf{c}_{\pi_{1}},\dots, \mathbf{c}_{\pi_{k}} \right]$ to be a $d \times k$ matrix. The following holds: 
\begin{equation*}
\left[\innp{\left(\mathbfcal{I}-\mathbfcal{V}\right)^{-1}\left(\pi_{n}\right)}{\nu_{i}}_\mathbb{H}\right]_{i=1,\dots,d}^{n=1,\dots,k}=\mathbf{G}^{\frac{1}{2}}\mathbf{D}\left(\sum_{l=0}^{\infty}\left(\mathbf{D}^{\top}\mathbf{D}\right)^{l}\right).
\end{equation*}
\end{lemma}
\noindent \textbf{Remark}: Note that since $\norm{\mathbfcal{V}}<1$, the real-valued sequence, $\left(\norm{\mathbfcal{V}^{l}}\right)_{l \in \mathbb{N}}$, converges to zero monotonically. As a result, we truncate $\sum_{l=0}^{\infty}\left(\mathbf{D}^{\top}\mathbf{D}\right)^{l}$ when $\norm{\left(\mathbf{D}^{\top}\mathbf{D}\right)^{l}}_{F}\approx 0$ for some $l \in \mathbb{N}$ where $\norm{\cdot}_{F}$ denotes the Frobenius norm of a matrix.

\subsection{FSSA R-Forecasting Algorithm}
As according to \eqref{eqn:rforecast}, our goal is to find the matrices that implement the effect of each $\mathbfcal{A}_{j}:\mathbb{H}_{d} \rightarrow \mathbb{H}_{d}$ for $j=1,\dots,L-1$.

\begin{theorem}\label{thm:rforecastimp}
Given that the conditions of Theorem \ref{thm:nunorm} hold, let 

$$\mathbf{A}_{j}=\mathbf{D}\left(\sum_{l=0}^{\infty}\left(\mathbf{D}^{\top}\mathbf{D}\right)^{l}\right)\mathbf{E}_{j}^{\top}\mathbf{G}^{\frac{1}{2}},\qquad\textrm{for}\ j=1,\dots, L-1,$$ 

\noindent where $\mathbf{E}_{j}=\mathbf{G}^{\frac{1}{2}}\left[\mathbf{c}_{\psi_{j,1}},\cdots, \mathbf{c}_{\psi_{j,k}}\right]$ be a $d \times k$ matrix. The corresponding coefficients of the function $\mathbfcal{A}_jf$ are given by $\mathbf{A}_j\mathbf{c}_f$.
\end{theorem}

\begin{corollary} 
The corresponding coefficients of the $g_i$'s given in \eqref{eqn:rforecast} can be written as

\[\mathbf{c}_{g_{i}}= \begin{cases} 
      \mathbf{c}_{\tilde{y}_{i}} & i=1,\dots,N \\
     \sum_{j=1}^{L-1}\mathbf{A}_{j}\mathbf{c}_{g_{i+j-L}} & i=N+1,\dots,N+M
   \end{cases}.\label{eqn:rforecastimp}\numberthis
\]
\end{corollary}





\subsection{FSSA V-Forecasting Algorithm}

In order to obtain the recipes for the FSSA V-forecasting algorithm, we first leverage theory developed in \cite{haghbin2019}. For some positive integer, $L$, we define $\mathbb{H}_{d}^{L}$ to be the space created from the Cartesian product of $L$ copies of $\mathbb{H}_{d}$. Now we define the quotient-remainder sequence

\begin{equation*}
j=\left(q_{j}-1\right)L+r_{j},\quad 1\leq q_{j} \leq d,\quad 1\leq r_{j} \leq L,
\end{equation*}

\noindent and from here we define the collection of linearly independent elements $\{\pmb{\phi}_{j}\}_{j=1}^{Ld}$ where $\pmb{\phi}_{j}\in \mathbb{H}_{d}^{L}$ is the zero function in all coordinates except for the $r_{j}^{\text{th}}$ which is $\nu_{q_{j}}$. We find that the basis, $\{\pmb{\phi}_{j}\}_{j=1}^{Ld}$, linearly spans $\mathbb{H}_{d}^{L}$. Like before,  any $\pmb{f} \in \mathbb{H}_{d}^{L}$ can be represented via its corresponding coefficient vector $\mathbf{c}_{\pmb{f}}=(c_{\pmb{f},1},\cdots,c_{\pmb{f},Ld})^\top \in \mathbb{R}^{Ld}$, where
$\pmb{f}=\sum_{i=1}^{Ld}{c}_{\pmb{f},i}\pmb{\phi}_{i}$. We denote the updated Gram matrix with $\mathbf{H}=\left[\innp{\pmb{\phi}_{i}}{\pmb{\phi}_{j}}_{\mathbb{H}^{L}}\right]_{i,j=1}^{Ld}$. Now, we define the new truncated quotient-remainder sequence as

\begin{equation*}\label{eqn:truncquorem}
i=\left(q_{i}-1\right)(L-1)+r_{i},\quad 1\leq q_{i} \leq d,\quad 1\leq r_{i} \leq L-1.
\end{equation*}

\noindent This allows us to define $\{\pmb{\phi}_{i}^{\nabla}\}_{i=1}^{\left(L-1\right)d}$ where $\pmb{\phi}_{i}^{\nabla} \in \mathbb{H}_{d}^{L-1}$ is zero in all coordinates except for the $r_{i}^{\text{th}}$ which is $\nu_{q_{i}}$, and $\mathbb{H}_{d}^{L-1}$ is the vector space formed from the Cartesian product of $L-1$ copies of $\mathbb{H}_{d}$. Similar to the non-truncated case, any $\pmb{f}^{\nabla}, \pmb{f}^{\Delta} \in \mathbb{H}_{d}^{L-1}$ can be represented via the corresponding coefficient vectors $\mathbf{c}_{\pmb{f}^{\nabla}}, \mathbf{c}_{\pmb{f}^{\Delta}} \in \mathbb{R}^{\left(L-1\right)d}$ associated to the basis functions $\{\pmb{\phi}_{i}^{\nabla}\}_{i=1}^{\left(L-1\right)d}$. Also, we define the respective Gram matrix, $\mathbf{H}^{\nabla}=\left[\innp{\pmb{\phi}_{i}^{\nabla}}{\pmb{\phi}_{j}^{\nabla}}_{\mathbb{H}^{L-1}}\right]_{i,j=1}^{\left(L-1\right)d}$. 

Now we must find the matrix that implements $\pmb{\Pi}:\mathbb{H}_{d}^{L-1} \rightarrow \mathbb{H}_{d}^{L-1}$. We again use ideas from \cite{haghbin2019} in the development of the following theory which gives the implementation of the FSSA V-forecasting algorithm.

\begin{theorem}\label{thm:vforecastimp}
Given that the conditions of Theorem \ref{thm:nunorm} hold, let 

$$\mathbf{P}=\left(\mathbf{F}\mathbf{F}^{\top}+\mathbf{F}\mathbf{D}^{\top}\mathbf{D}\sum_{l=0}^{\infty}\left(\mathbf{D}^{\top}\mathbf{D}\right)^{l}\mathbf{F}^{\top}\right)\left(\mathbf{H}^{\nabla}\right)^{\frac{1}{2}},$$ 

\noindent where $\mathbf{F}=\left(\mathbf{H}^{\nabla}\right)^{\frac{1}{2}}\left[\mathbf{c}_{\pmb{\psi}_{1}^{\nabla}},\cdots, \mathbf{c}_{\pmb{\psi}_{k}^{\nabla}}\right]$ be a $(L-1)d \times k$ matrix. For any $\pmb{f}\in\mathbb{H}^{(L-1)}_d$, the corresponding coefficients of the function $\pmb{\Pi}\pmb{f}$ are given by $\mathbf{P}\mathbf{c}_{\pmb{f}}$.
\end{theorem}

\begin{theorem}\label{thm:fullvforecastimp}
For any $\pmb{x}\in\mathbb{L}$, let the linear operator $\mathbfcal{Q}$ to be the one given in \eqref{eqn:Q}. The corresponding coefficients of $\mathbfcal{Q}\pmb{x}$ is given by 
\begin{equation*}\label{eqn:vforecastimp}
\mathbf{c}_{\mathbfcal{Q}}(\pmb{x})= \begin{pmatrix}\mathbf{P}\mathbf{c}_{\pmb{x}^{\Delta}} \\ \sum_{j=1}^{L-1}\mathbf{A}_{j}\mathbf{c}_{x_{j}^{\Delta}}\end{pmatrix}
\end{equation*}
\end{theorem}

\begin{corollary}

The corresponding coefficients of the $\pmb{w}_j$'s given in \eqref{eqn:vforecast} can be written as

\[\mathbf{c}_{\pmb{w}_{j}}= \begin{cases} 
      \mathbf{c}_{\pmb{x}_{j}^{\mathfrak{s}}} & j=1,\dots, K \\
      \mathbf{c}_{\mathbfcal{Q}}(\pmb{w}_{j-1}) & j=K+1, \dots, K+M
   \end{cases}.\label{eqn:vforecastimp}\numberthis
\]

\end{corollary}

\section{Numerical Studies}\label{sec:num}

In this section, we offer a simulation study and real data examples that showcase the advantage of our novel methods in forecasting periodic FTS. In the simulation study, we compare our approaches with the functional seasonal naive method and we also compare to the popular FTS forecasting technique of \cite{hyndman2007}, which we now call the competing method. We incorporate either seasonality, increasing trend, or both components in various simulation setups. We further illustrate the superior performance in forecasting of a periodic FTS with a real data study where we compare our techniques with the competing method when the optimal parameters are selected for each algorithm applied to highly periodic call center data, moderately periodic remote sensing data, and mortality rate data that has no periodic components. We note that when we state that we are using $k$ eigentriples in forecasts, that means we use $k$ left singular functions in the prediction.

\subsection{Simulation Study}\label{subsec:sim}

For the simulation study, we use a setup that is similar to that seen in \cite{haghbin2019} and \cite{trinka2020} where varying cases will generate FTS that have only periodicty or only trend or both components. This particular setup utilizes FTS of lengths $N=100, 200$ that are observed on $n = 100$ fixed, equidistant discrete points on the unit interval from the following model:

\begin{equation*}
Y_{t}\left(s_{i}\right)=m_{t}\left(s_{i}\right)+X_{t}\left(s_{i}\right),\quad s_{i}\in\left[0,1\right],i=1,\dots,n,\text{ and } t=1,\dots,N.
\end{equation*}

\noindent We use a B-spline basis with 15 degrees of freedom to smooth the discrete samplings of functional curves. We have that $m_{t}\left(s\right)$ is the true underlying signal to be extracted and predicted where

\begin{equation*}
m_{t}\left(s\right)=\kappa t+e^{s^{2}}\cos\left(2\pi\omega t\right)+\cos\left(4\pi s\right)\sin\left(2\pi\omega t\right), 
\end{equation*}

\noindent and we allow the trend coefficient, $\kappa$, to take on values of $0$ and $0.02$ while the frequency, $\omega$, takes on the values of $0$ or $0.20$ in varying setups. We also have that $X_{t}\left(s\right)$ is a stochastic term that follows a functional autoregressive model of order 1, FAR(1), defined by

\begin{equation*}
X_{t}\left(s\right) = \Psi X_{t-1}\left(s\right) +\epsilon_{t}\left(s\right),
\end{equation*}

\noindent where $\Psi$ is an integral operator with a parabolic kernel as follows

\begin{equation*}
\psi\left(s,u\right) = \gamma_{0}\left(2-\left(2s-1\right)^{2}-\left(2u-1\right)^{2}\right).
\end{equation*}

\noindent We choose $\gamma_{0}$ such that the Hilbert-Schmidt norm defined by

\begin{equation*}
\norm{\Psi}_{\mathcal{S}}^{2}=\int_{0}^{1}\int_{0}^{1}\abs{\psi\left(s,u\right)}^{2}dsdu,
\end{equation*}

\noindent takes on values of $0.25$, $0.60$, $0.90$, and $0.95$. We also consider the terms $\epsilon_{t}\left(s\right)$ to be independent trajectories of standard Brownian motion over $\left[0,1\right]$. We will be comparing our FSSA recurrent forecast (RFSSA) to our FSSA vector forecast (VFSSA) with varying lags, $L$, of $10$ and $20$ and we compare both of these methods to the functional seasonal naive method (SNM) and the competing method (H \& U). We choose the cases for the lag parameter in accordance with SSA literature where $L$ is often chosen to be a multiple of the periodicity in the data, see \cite{golyandina2001}, \cite{haghbin2019}, and \cite{trinka2020}. In addition, we choose the optimal number of eigentriples to perform reconstructions and forecasts in FSSA-based models for each combination of $\kappa$ and $\omega$ (two when $\kappa=0$, $\omega=0.20$, three when $\kappa=0.02$, $\omega=0.20$, and one when $\kappa=0.02$, $\omega=0$). We have that the maximum number of left singular functions chosen in the FSSA-based predictions is three (when $\kappa=0.02$ and $\omega=0.20$) and due to this, we choose three functional principal components to perform all forecasts in the method of \cite{hyndman2007} for the simulations.

We expect that since our methods incorporate periodicity into the left singular functions that are used in the forecast, our techniques will perform better than the competing approach in predicting periodic true signals. To perform the comparison, we leverage a rolling forecast where the size of our training set, $O$, takes on values of $60$ and $80$. We denote the forecast of the $t^{\text{th}}$ function evaluated at $s_{i} \in \left[0,1\right]$ with $\hat{Y}_{t}\left(s_{i}\right)$, which allows us to define the root mean square error of

\begin{equation*}
RMSE=\sqrt{\frac{1}{\left(N-O\right) \times n}\sum_{t=1}^{N-O}\sum_{i=1}^{n}\left(\hat{Y}_{t+O}\left(s_{i}\right)-m_{t+O}\left(s_{i}\right)\right)^{2}}. 
\end{equation*}

\begin{figure}[H]
	\centering
	\includegraphics[page=1,width=\textwidth]{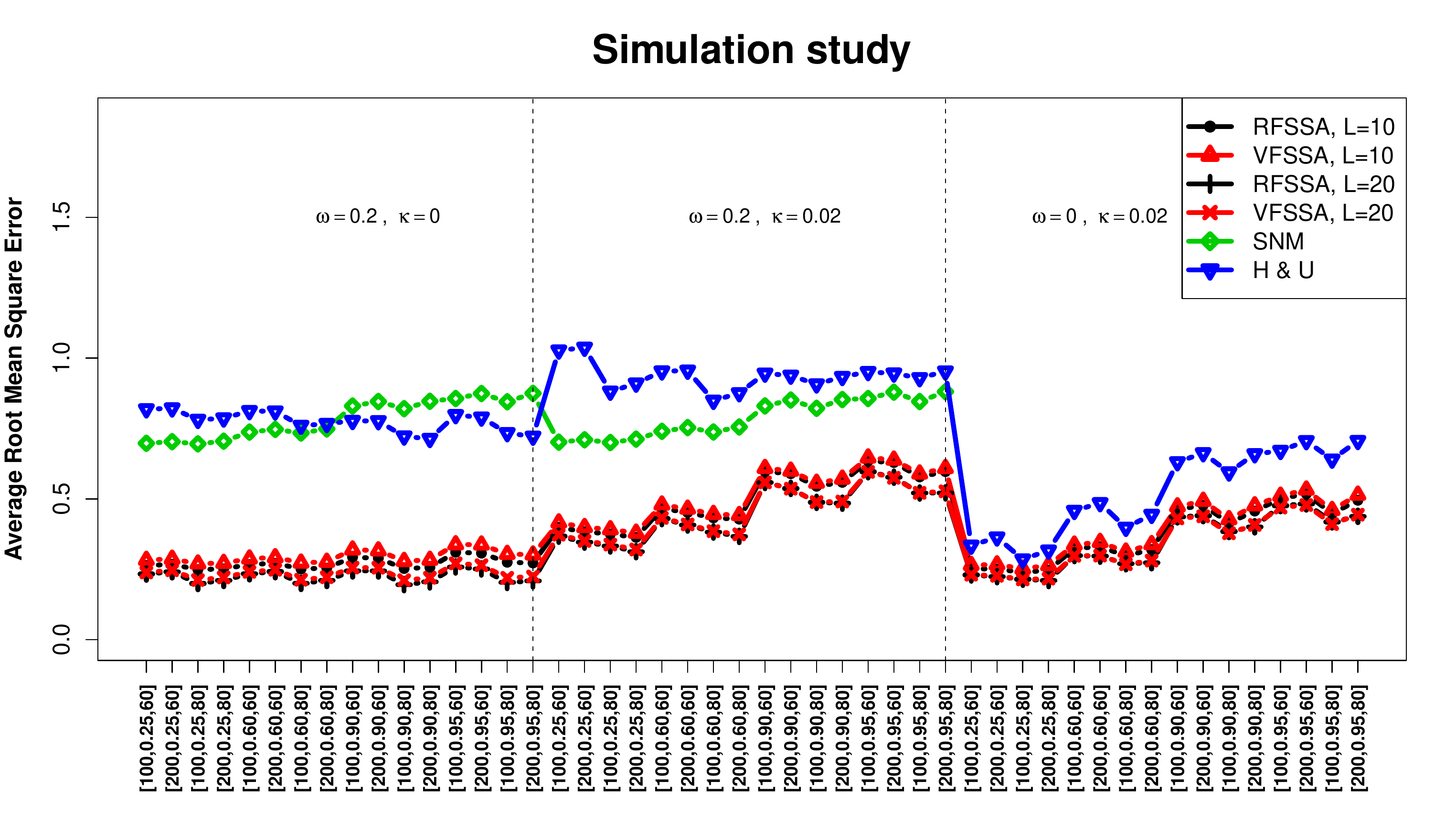}
	\caption{Simulation study where horizontal axis entries have form of $\left[N, \norm{\Psi}_{\mathcal{S}}^{2}, O\right]$.}
		\label{fig:sim}
\end{figure}

\noindent From here, we replicate the result of each simulation setup combination $100$ times and report the mean of the RMSE's in Figure \ref{fig:sim}.

\noindent We find that our novel approaches consistently outperform the SNM and the competing methodology for all setups. These results further illustrate the superior performance of our novel methodologies in predicting the true underlying signal when periodicity is present in the data and how our methods are still competitive for data that have no oscillatory elements. For a visuanimation that displays a sample of the simulation setups and some of the graphical results, see the supplementary material.

We note that principal components that account for small variation in the data can still play a big role in improving predictions in principal component regression, see \cite{joliffe1982} and references therein. As a result, we may have to use more than three functional principal components when applying the method of \cite{hyndman2007} in FTS forecasting however, the fact that our methods outperform the competing approach with such few elements used is a bonus. Now in the real data study, we showcase that our methodologies still outperform the FPCA-based technique of \cite{hyndman2007} even when the optimal number of functional principal components are chosen for the competing algorithm.

\subsection{Real Data Study}\label{subsec:real}
\begin{figure}[H]
	\centering
	\includegraphics[page=1,width=.33\textwidth]{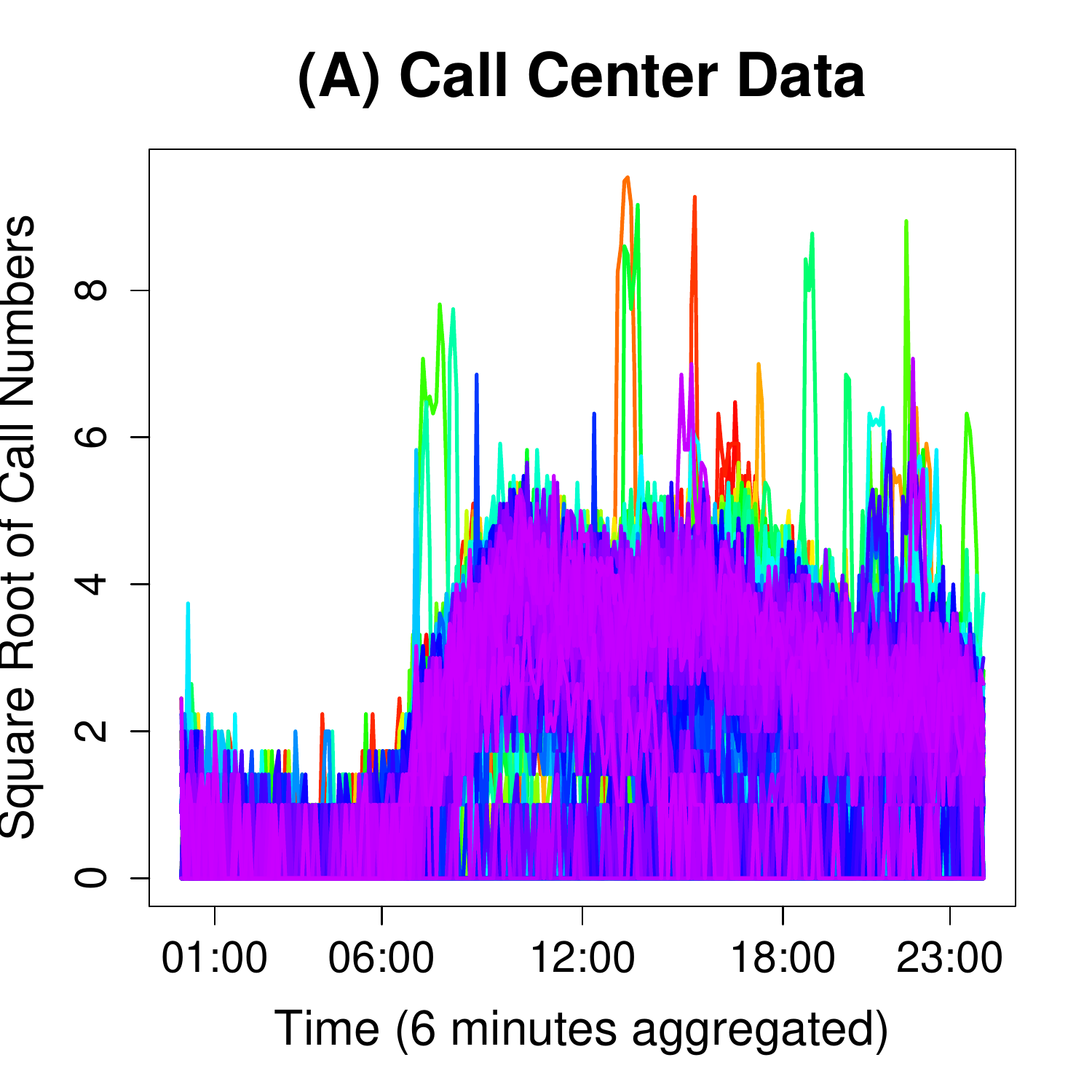}
	\includegraphics[page=1,width=.33\textwidth]{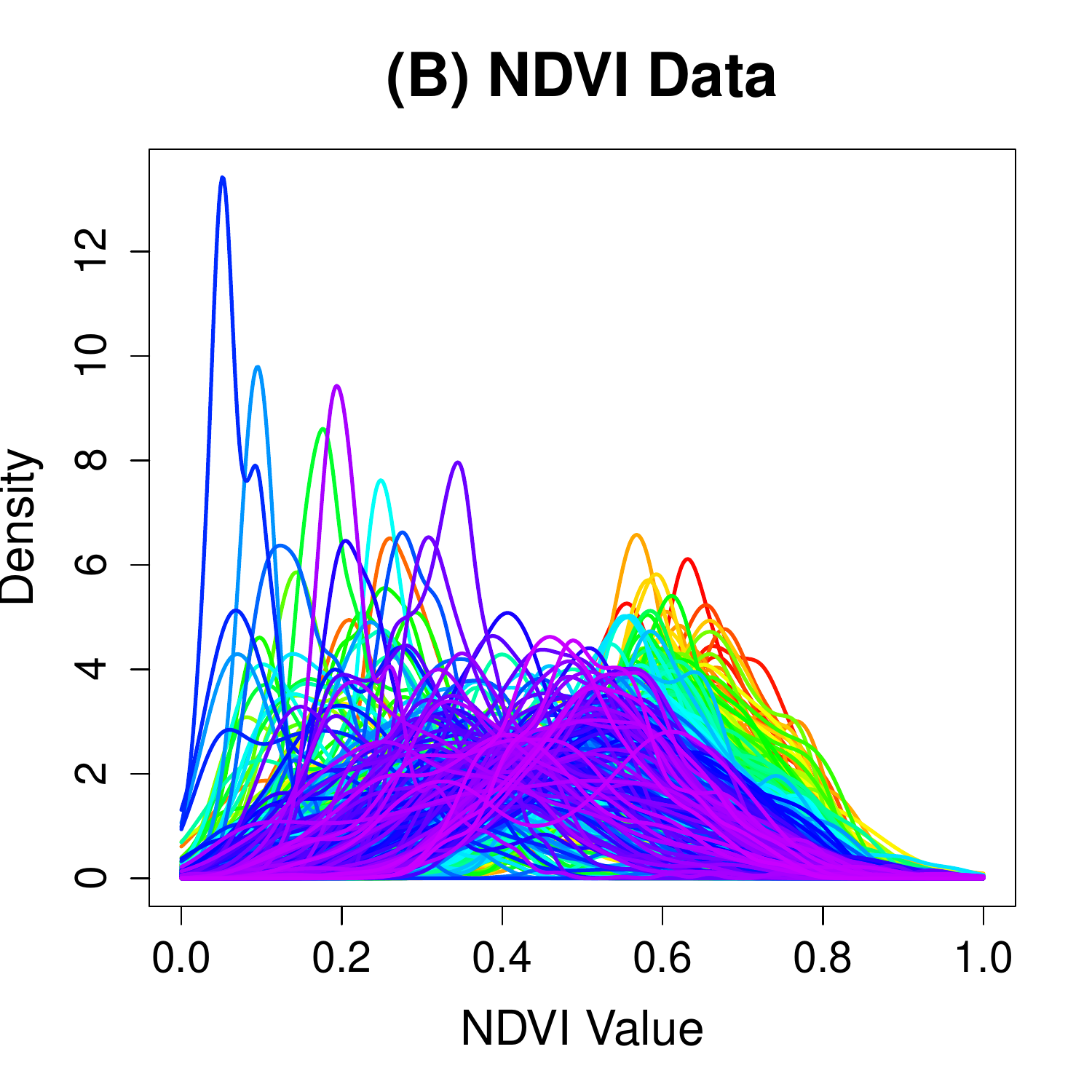}
	\includegraphics[page=1,width=.32\textwidth]{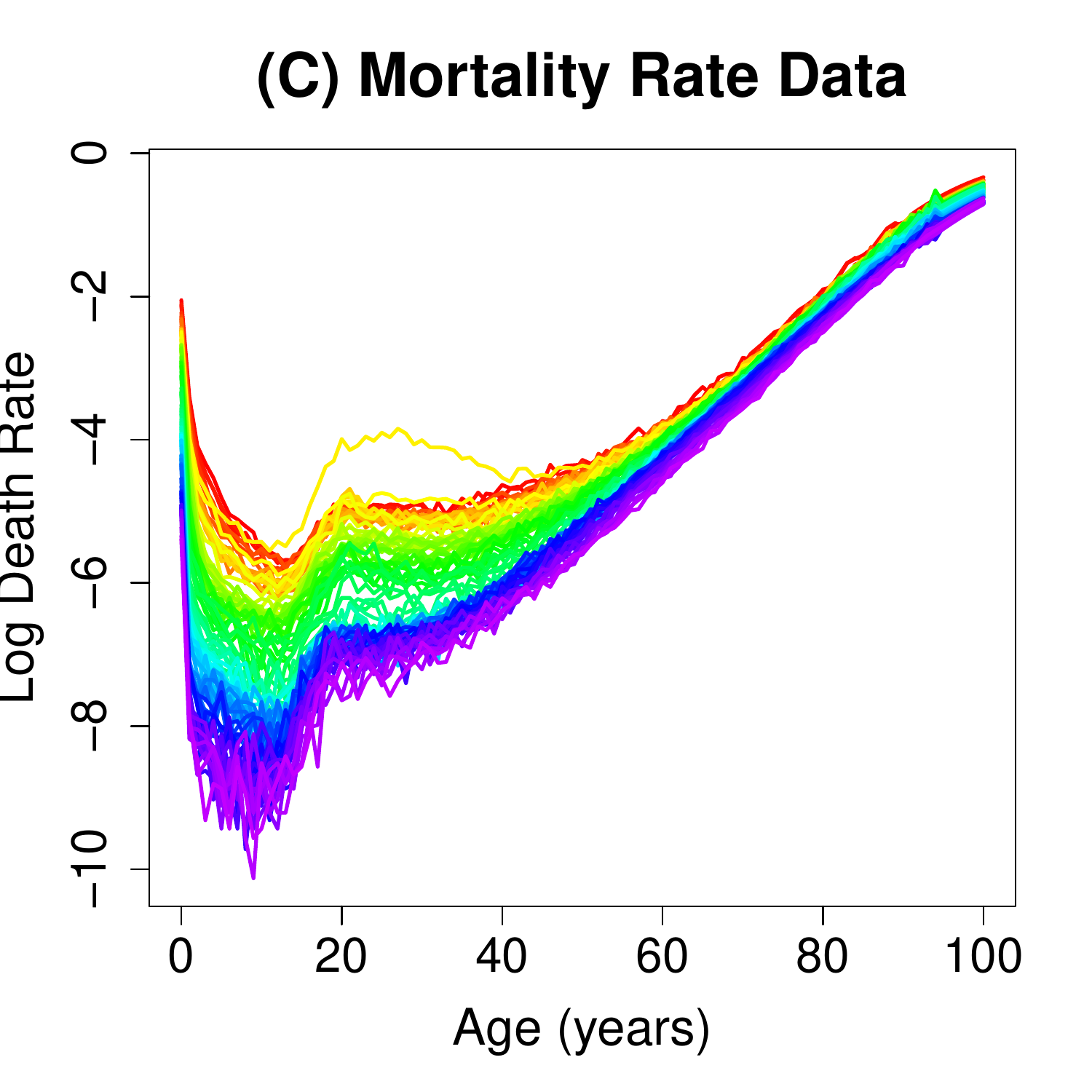}
	\caption{(A): Call center data, (B): NDVI densities, (C): Swedish mortality rate data. Warm colors, such as red and yellow, are indcative of curves observed on earlier dates while cool colors, such as blue and purple, are indicative of curves observed on later dates. Figures generated using the \pkg{rainbow} package \citep{rainbowpackage}.}
		\label{fig:real_data}
\end{figure}

We first introduce three real FTS each with varying levels of periodicity and trend being present. The first FTS is the call center data analyzed in Section \ref{sec:int} and we show this data again in Figure \ref{fig:real_data}(A). The next is NDVI remote sensing data where NDVI images can be used to remotely track changes in vegetation \citep{Panuju2012, Tuck2014, Lambin1999}. NDVI values closer to zero are indicative of less vegetation being present in a part of an image while values closer to one are indicative of more vegetation being present. From NDVI images taken of Jambi, Indonesia between February 18, 2000 and July 28, 2019 in $16$ day increments, we estimate $448$ densities of NDVI values using Silverman's rule of thumb \citep{silverman1986}, which can be seen in Figure \ref{fig:real_data}(B). We also note that an example of an NDVI image and associated density can be found in the supplementary material. The final is a dataset of $97$ functions representative of mortality rate data of Swedish males between ages $0$ and $100$ for years 1899 to 1995 seen in Figure \ref{fig:real_data}(C) \citep{mortalitydata}.

Using Figure \ref{fig:real_data}, we can infer existence of periodic and trend behaviors present in each data set. Figure \ref{fig:real_data}(A) clearly shows periodic behavior with no trend as all warmer colors are covered with cooler colors. This strong, weekly periodicity for the call center data is also uncovered in \cite{haghbin2019} and a summary of the results are given in the supplementary materials. Plot (B) shows a slight decrease in NDVI values over time since densities observed in earlier dates (warmer colors) cluster around NDVI values of $0.6$ while densities observed in later dates (cooler colors) cluster around $0.5$. The existence of such a decreasing trend and an annual periodic behavior are confirmed by the work of \cite{haghbin2019} and we offer an outline of these results in the supplement. Finally, we see a strong decreasing trend with no periodicity in the mortality rate data in Figure \ref{fig:real_data}(C). Due to the fact that our novel methods appear to perform best when a FTS is periodic as shown in the simulation study, we expect our novel methods to perform better than the competing method for the highly periodic call center data and we expect a closer competition for the NDVI densities and mortality rate data.
 
 As according to the work of \cite{haghbin2019}, for the FSSA-based prediction algorithms applied to call center data, we select $L=28$ and the first seven eigentriples to construct our deterministic FTS in reconstruction and to perform the forecast. Also according to \cite{haghbin2019}, for the FSSA-based algorithms applied to NDVI data, we select $L=45$ and the first four eigentriples to build the FTS and perform the forecast. Since there are no periodic elements present in the mortality rate data, we cannot use the standard rule of leveraging a lag that is a multiple of the periodicity in the FTS for the FSSA-based algorithms. In addition, we must select the best number of eigentriples to reconstruct the deterministic signal and to do the prediction given some lag $L$ for the mortality rate data. We also must select the optimal number of functional principal components to use in the forecasts while applying the method of \cite{hyndman2007} to each of the three data sets. To achieve these goals, we look towards cross-validation via a rolling forecast. We first define prediction root mean square error by

\begin{equation}\label{eqn:trmse}
prRMSE=\sqrt{\frac{1}{\left(N-O\right) \times n}\sum_{t=1}^{N-O}\sum_{i=1}^{n}\left(\hat{Y}_{t+O}\left(s_{i}\right)-Y_{t+O}\left(s_{i}\right)\right)^{2}}
\end{equation}

\noindent where $\hat{Y}_{t+O}$ is the prediction of observation $Y_{t+O}$, $s_{i}$ is a point in the domain, $n$ are the number of sampling points, $N$ is the length of the FTS, and $O$ is the training set size. We perform rolling forecasts with training set sizes of $308$, $403$, and $50$ for the call center, NDVI, and mortality data respectively (giving us testing set sizes of $57$, $45$, and $47$ respectively), and we estimate $prRMSE$ as according to \eqref{eqn:trmse} for varying choices of parameters. We find that a lag of $24$ and using the first eigentriple to perform reconstruction and forecasting in the FSSA-based algorithms applied to the mortality rate data minimize \eqref{eqn:trmse} for that data set. In terms of choosing the optimal number of functional principal components to perform the forecast in the method of \cite{hyndman2007}, we find choosing the first $33$ functional principal components, the first eight functional principal components, and the first three functional principal components for the call center, NDVI, and mortality rate data respectively minimize \eqref{eqn:trmse}. We report the prediction root mean square errors for each optimal model in Table \ref{tbl:trmse}.
\begin{table}[H]
\centering
\begin{tabular}{|l|l|l|l|}
\hline
Methodology                                   & Call Center & NDVI  & Mortality \\ \hline
FSSA V-Forecasting                                 & 0.567       & 0.824 & 0.227    \\ \hline
FSSA R-Forecasting                                 & 0.569       & 0.827 &  0.236     \\ \hline
Method of \cite{hyndman2007} & 0.855      & 0.818 & 0.143     \\ \hline
\end{tabular}
\caption{$prRMSE$'s of FSSA R-forecasting, FSSA V-forecasting, and the competing method of \cite{hyndman2007} for various datasets}
\label{tbl:trmse}
\end{table}

\noindent As seen in Table \ref{tbl:trmse}, our novel methodologies outperform for the highly periodic call center data. The three methods are comparable for the NDVI data which has a mixture of trend and periodic components. Finally, we see that the competing method does in fact perform better for the mortality rate data which contains no periodic components and only a decreasing trend.

\section{Discussion}\label{sec:discussion}
In this work, we developed efficient nonparametric FTS forecasting techniques that include periodicity into the basis elements. We compared our approaches with the competing method of \cite{hyndman2007} and found our techniques are superior in forecasting periodic FTS since the FSSA routine captures periodic behavior in the basis elements. As a result researchers can use our methodologies in order to obtain accurate and informative predictions of periodic stochastic processes.

\bibliographystyle{apalike}
\bibliography{Mybib}

\begin{thebibliography}{}

\bibitem[Beyaztas and Shang, 2019]{beyaztas2019}
Beyaztas, U. and Shang, H.~L. (2019).
\newblock Forecasting functional time series using weighted likelihood
  methodology.
\newblock {\em Journal of Statistical Computation and Simulation},
  89(16):3046--3060.

\bibitem[Golyandina et~al., 2015]{golyandina2015}
Golyandina, N., Korobeynikov, A., Shlemov, A., and Usevich, K. (2015).
\newblock Multivariate and 2d extensions of singular spectrum analysis with the
  rssa package.
\newblock {\em Journal of Statistical Software, Articles}, 67(2):1--78.

\bibitem[Golyandina et~al., 2001]{golyandina2001}
Golyandina, N., Nekrutkin, V., and Zhigljavsky, A.~A. (2001).
\newblock {\em {Analysis of time series structure: SSA and related
  techniques}}.
\newblock Chapman and Hall/CRC.

\bibitem[Golyandina and Zhigljavsky, 2013]{golyandina2013}
Golyandina, N. and Zhigljavsky, A. (2013).
\newblock {\em {Singular spectrum analysis for time series}}.
\newblock Springer Science \& Business Media.

\bibitem[{Gonz\'{a}lez} et~al., 2018]{gonzalez2018}
{Gonz\'{a}lez}, J.~P., {Mu\~{n}oz San Roque}, A. M.~S., and {P\'{e}rez}, E.~A.
  (2018).
\newblock Forecasting functional time series with a new hilbertian armax model:
  Application to electricity price forecasting.
\newblock {\em IEEE Transactions on Power Systems}, 33(1):545--556.

\bibitem[{Haghbin} et~al., 2020a]{haghbin2019}
{Haghbin}, H., {Morteza Najibi}, S., {Mahmoudvand}, R., {Trinka}, J., and
  {Maadooliat}, M. (accepted 2020a).
\newblock {Functional Singular Spectrum Analysis}.
\newblock {\em Stat}.
\newblock Retreived from \url{https://arxiv.org/abs/1906.05232}.

\bibitem[{Haghbin} et~al., 2020b]{haghbin2019supp}
{Haghbin}, H., {Morteza Najibi}, S., {Mahmoudvand}, R., {Trinka}, J., and
  {Maadooliat}, M. (accepted 2020b).
\newblock {Functional Singular Spectrum Analysis Supplementary Material}.
\newblock {\em Stat}.
\newblock Retreived from \url{https://arxiv.org/abs/1906.05232}.

\bibitem[Happ and Greven, 2018]{happ2018}
Happ, C. and Greven, S. (2018).
\newblock Multivariate functional principal component analysis for data
  observed on different (dimensional) domains.
\newblock {\em Journal of the American Statistical Association}, 113(522):649
  -- 659.

\bibitem[Hassani and Mahmoudvand, 2013]{hassani2013}
Hassani, H. and Mahmoudvand, R. (2013).
\newblock Multivariate singular spectrum analysis: A general view and new
  vector forecasting approach.
\newblock {\em International Journal of Energy and Statistics}, 01(01):55--83.

\bibitem[Huang et~al., 2008]{huang2008functional}
Huang, J.~Z., Shen, H., Buja, A., et~al. (2008).
\newblock Functional principal components analysis via penalized rank one
  approximation.
\newblock {\em Electronic Journal of Statistics}, 2:678--695.

\bibitem[Hyndman and Shang, 2009]{hyndman2009}
Hyndman, R. and Shang, H.~L. (2009).
\newblock Functional time series forecasting.
\newblock {\em Journal of the Korean Statistical Society}, 38:199--211.

\bibitem[Hyndman and Shang, 2020]{ftsapackage}
Hyndman, R. and Shang, H.~L. (2020).
\newblock {\em {ftsa: Functional Time Series Analysis}}.
\newblock R package version 5.9.0.

\bibitem[Hyndman and Ullah, 2007]{hyndman2007}
Hyndman, R. and Ullah, S. (2007).
\newblock Robust forecasting of mortality and fertility rates: A functional
  data approach.
\newblock {\em Computational Statistics \& Data Analysis}, 51:4942--4956.

\bibitem[Jeng-Min et~al., 2014]{chiou2014}
Jeng-Min, C., Yu-Ting, C., and Ya-Fang, Y. (2014).
\newblock Multivariate functional principal component analysis: A normalization
  approach.
\newblock {\em Statistica Sinica}, 24(4):1571.

\bibitem[Jolliffe, 1982]{joliffe1982}
Jolliffe, I.~T. (1982).
\newblock A note on the use of principal components in regression.
\newblock {\em Journal of the Royal Statistical Society: Series C (Applied
  Statistics)}, 31(3):300--303.

\bibitem[Lambin, 1999]{Lambin1999}
Lambin, E.~F. (1999).
\newblock {Monitoring forest degradation in tropical regions by remote sensing:
  Some methodological issues}.
\newblock {\em Global Ecology and Biogeography}, 8(3-4):191--198.

\bibitem[Maadooliat et~al., 2015]{maadooliat2015integrating}
Maadooliat, M., Huang, J.~Z., and Hu, J. (2015).
\newblock Integrating data transformation in principal components analysis.
\newblock {\em Journal of Computational and Graphical Statistics},
  24(1):84--103.

\bibitem[Panuju and Trisasongko, 2012]{Panuju2012}
Panuju, D.~R. and Trisasongko, B.~H. (2012).
\newblock Seasonal pattern of vegetative cover from {NDVI} time-series.
\newblock {\em Tropical Forests}, page 255.

\bibitem[Ramsay and Silverman, 2005]{ramsay2005}
Ramsay, J.~O. and Silverman, B.~W. (2005).
\newblock {\em Functional data analysis.}
\newblock Springer series in statistics. Springer.

\bibitem[Shang, 2013]{shang2013}
Shang, H.~L. (2013).
\newblock Functional time series approach for forecasting very short-term
  electricity demand.
\newblock {\em Journal of Applied Statistics}, 40(1):152--168.

\bibitem[Shang, 2019]{shang2019}
Shang, H.~L. (2019).
\newblock A robust functional time series forecasting method.
\newblock {\em Journal of Statistical Computation and Simulation}, 89:795--814.

\bibitem[Shang and Hyndman, 2019]{rainbowpackage}
Shang, H.~L. and Hyndman, R. (2019).
\newblock {\em {rainbow: Bagplots, Boxplots and Rainbow Plots for Functional
  Data}}.
\newblock R package version 3.6.0.

\bibitem[Shen and Huang, 2005]{shen2005analysis}
Shen, H. and Huang, J.~Z. (2005).
\newblock {Analysis of call centre arrival data using singular value
  decomposition}.
\newblock {\em Applied Stochastic Models in Business and Industry},
  21(3):251--263.

\bibitem[Silverman, 1986]{silverman1986}
Silverman, B. (1986).
\newblock {\em Density estimation for statistics and data analysis}.
\newblock Chapman \& Hall, London.

\bibitem[Trinka et~al., 2020]{trinka2020}
Trinka, J., Haghbin, H., and Maadooliat, M. (2020).
\newblock Multivariate functional singular spectrum analysis over different
  dimensional domains.

\bibitem[Tuck et~al., 2014]{Tuck2014}
Tuck, S.~L., Phillips, H.~R., Hintzen, R.~E., Scharlemann, J.~P., Purvis, A.,
  and Hudson, L.~N. (2014).
\newblock {MODISTools -- downloading and processing MODIS remotely sensed data
  in R}.
\newblock {\em Ecology and Evolution}, 4(24):4658--4668.

\bibitem[{University of California at Berkely (USA) and Max Planck Institue for
  Demographic Research}, 2020]{mortalitydata}
{University of California at Berkely (USA) and Max Planck Institue for
  Demographic Research} (2020).
\newblock Human mortality database.
\newblock Data retrieved from \url{https://www.mortality.org/}.

\bibitem[{Wagner-Muns} et~al., 2018]{wagner2018}
{Wagner-Muns}, I.~M., {Guardiola}, I.~G., {Samaranayke}, V.~A., and {Kayani},
  W.~I. (2018).
\newblock A functional data analysis approach to traffic volume forecasting.
\newblock {\em IEEE Transactions on Intelligent Transportation Systems},
  19(3):878--888.

\end{thebibliography}

\newpage
\setcounter{page}{1}
\renewcommand{\thefigure}{S\arabic{figure}}
\renewcommand{\thetable}{S\arabic{table}}
\renewcommand{\theequation}{S\arabic{equation}}
\renewcommand{\thesection}{S\arabic{section}}
\setcounter{figure}{0}   
\setcounter{table}{0}   
\setcounter{equation}{0}
\setcounter{section}{0}   
\section*{Supplementary Materials}

The following contains supplementary materials to the manuscript in order to provide further clarification. We first offer all proofs of theory presented in the manuscript. Next, we give a visuanimation which shows a subset of simulation setups leveraged in the simulation study of the manuscript for futher visualization. Then we offer FSSA analysis performed on call center and NDVI data to futher illustrate the existence or lack of periodic or trend behaviors in these datasets. Then we give a review of the competing method of \cite{hyndman2007}.

\subsection*{Proofs}

\begin{proof}[Proof of Thm.]\ref{thm:nunorm}

Fix $\pmb{e}_{x}=\left(\underline{0},\dots,\underline{0},x\right)$ to be a point in $\mathbb{E}$. Since $\{\pmb{\psi}_{j}\}_{j=1}^{k}$ is an orthonormal basis that linearly spans $\mathbb{L}$, then we may express the orthogonal projection of $\pmb{e}_{x}$ onto $\mathbb{L}$ as

\begin{equation*}
\pmb{p}=\sum_{j=1}^{k}\innp{\pmb{e}_{x}}{\pmb{\psi}_{j}}_{\mathbb{H}^{L}}\pmb{\psi}_{j}.
\end{equation*}

\noindent Notice that $\innp{\pmb{e}_{x}}{\pmb{\psi}_{j}}_{\mathbb{H}^{L}}=\innp{x}{\pi_{j}}_{\mathbb{H}}$ for $j=1,\dots,k$, as such we find that

\begin{equation*}
\pmb{p}=\begin{pmatrix}\sum_{j=1}^{k} \innp{x}{\pi_{j}}_{\mathbb{H}}\psi_{1,j} \\ \vdots \\ \sum_{j=1}^{k} \innp{x}{\pi_{j}}_{\mathbb{H}}\psi_{L-1,j} \\ \sum_{j=1}^{k} \innp{x}{\pi_{j}}_{\mathbb{H}}\pi_{j} \end{pmatrix}=\begin{pmatrix}\sum_{j=1}^{k} \innp{x}{\pi_{j}}_{\mathbb{H}}\psi_{1,j} \\ \vdots \\ \sum_{j=1}^{k} \innp{x}{\pi_{j}}_{\mathbb{H}}\psi_{L-1,j} \\ \mathbfcal{V}\left(x\right) \end{pmatrix}.
\end{equation*}

\noindent We let $p_{i}=\sum_{j=1}^{k}\innp{x}{\pi_{j}}_{\mathbb{H}}\psi_{i,j}$ for $i=1,\dots,L-1$, and we find that

\begin{equation*}
\norm{\pmb{p}}_{\mathbb{H}^{L}}^{2}=\sum_{i=1}^{L-1}\norm{p_{i}}_{\mathbb{H}}^{2}+\norm{\mathbfcal{V}\left(x\right)}_{\mathbb{H}}^{2}.
\end{equation*}

Since $\pmb{p}$ is an orthogonal projection of $\pmb{e}_{x}$ onto $\mathbb{L}$, and the projection is non-zero, then, we have that $\norm{\pmb{p}}_{\mathbb{H}^{L}}^{2}=\sum_{i=1}^{L-1}\norm{p_{i}}_{\mathbb{H}}^{2}+\norm{\mathbfcal{V}\left(x\right)}_{\mathbb{H}}^{2} \leq \norm{\pmb{e}_{x}}_{\mathbb{H}^{L}}^{2}$. This implies that

\begin{equation*}
\norm{\mathbfcal{V}\left(x\right)}_{\mathbb{H}}^{2}\leq \norm{\pmb{e}_{x}}_{\mathbb{H}^{L}}^{2}=\norm{x}_{\mathbb{H}}^{2}.
\end{equation*}

\noindent Since $\norm{\mathbfcal{V}\left(x\right)}_{\mathbb{H}}^{2} \leq \norm{x}_{\mathbb{H}}^{2}$, we have that $\mathbfcal{V}$ is in fact a contraction and $\norm{\mathcal{V}} < 1$.
\end{proof}

\begin{proof}[Proof of Thm.]\ref{thm:rforecast}

Recall that $\{\pmb{\psi}_{n}^{\nabla}\}_{n=1}^{k}$ spans $\mathbb{L}^{\nabla}$, then for each $\pmb{v} \in \mathbb{L}^{\nabla}$ there exists a unique collection of coefficients $\{h_{n}\}_{n=1}^{k}$ such that

\begin{equation*}
\pmb{v}=\sum_{n=1}^{k}h_{n}\pmb{\psi}_{n}^{\nabla}.
\end{equation*}

\noindent Also recall that $\{\pmb{\psi}_{n}\}_{n=1}^{k}$ spans $\mathbb{L}$ and as such, any $\pmb{y} \in \mathbb{L}$ can be expressed as

\begin{equation*}
\pmb{y}=\sum_{n=1}^{k}h_{n}\pmb{\psi}_{n}.
\end{equation*}

\noindent From this, we see that $\pmb{y}^{\nabla}=\sum_{n=1}^{k}h_{n}\pmb{\psi}_{n}^{\nabla}=\pmb{v}$. As for the uniqueness of $\pmb{y}$, let $\pmb{y}_{1}$, $\pmb{y}_{2} \in \mathbb{L}$ and let $\pmb{y}_{1}^{\nabla}=\pmb{y}_{2}^{\nabla}=\pmb{v}$, then $\pmb{y}_{1}-\pmb{y}_{2} \in \mathbb{L}$ and is proportional to $\pmb{e}_{x}$ which is a contradiction and implies that $\pmb{y}_{1}=\pmb{y}_{2}$. Thus, we see that $\pmb{y}$ must be unique.

Now we show that for some functional $L$-lagged vector $\pmb{y}=\left(y_{1},\dots,y_{L}\right)^{\top} \in \mathbb{L}$, the last component, $y_{L}$, is a linear combination of the previous $L-1$ components. Let $\pmb{v}=\left(y_{1},\dots,y_{L-1}\right)^{\top} \in \mathbb{L}^{\nabla}$ and notice that

\begin{equation}\label{eqn:seplast}
\left(y_{1},\dots,y_{L-1},\underline{0}\right)^{\top}+\left(\underline{0},\dots,\underline{0},y_{L}\right)^{\top}=\sum_{n=1}^{k}h_{n}\pmb{\psi}_{n}.
\end{equation}

\noindent Taking the inner product of the left and right-hand sides of \eqref{eqn:seplast} with each $\pmb{\psi}_{n}$ gives us

\begin{equation}\label{eqn:coefs}
h_{n}=\innp{\pmb{v}}{\pmb{\psi}_{n}^{\nabla}}_{\mathbb{H}^{L-1}}+\innp{y_{L}}{\pi_{n}}_{\mathbb{H}},\quad n=1,\dots,k.
\end{equation}

\noindent Notice that we may express $y_{L}$ as 

\begin{equation}\label{eqn:lastcompexp}
y_{L}=\sum_{n=1}^{k} h_{n}\pi_{n}
\end{equation}

\noindent and now we substitute in the right-hand side of \eqref{eqn:coefs} for $h_{n}$ in \eqref{eqn:lastcompexp} to obtain

\begin{equation*}
y_{L}=\sum_{n=1}^{k}\innp{\pmb{\psi}_{n}^{\nabla}}{\pmb{v}}_{\mathbb{H}^{L-1}}\pi_{n}+\mathbfcal{V}\left(y_{L}\right).
\end{equation*}

\noindent We subtract $\mathbfcal{V}\left(y_{L}\right)$ to the left-hand side of the equation and factor out $y_{L}$. We assume that $\mathbb{E} \cap \mathbb{L}=\emptyset$ such that $\left(\mathbfcal{I}-\mathbfcal{V}\right)^{-1}$ exists and we find that

\begin{equation*}
y_{L}=\sum_{n=1}^{k}\innp{\pmb{\psi}_{n}^{\nabla}}{\pmb{v}}_{\mathbb{H}^{L-1}}\left(\mathbfcal{I}-\mathbfcal{V}\right)^{-1}\pi_{n}.
\end{equation*}

\noindent Recall that $\innp{\pmb{\psi}_{n}^{\nabla}}{\pmb{v}}_{\mathbb{H}^{L-1}}=\sum_{j=1}^{L-1}\innp{\psi_{j,n}}{y_{j}}_{\mathbb{H}}$ and thus

\begin{align*}
y_{L}=\sum_{n=1}^{k}\sum_{j=1}^{L-1}\innp{\psi_{j,n}}{y_{j}}_{\mathbb{H}}\left(\mathbfcal{I}-\mathbfcal{V}\right)^{-1}\pi_{n}&=\sum_{j=1}^{L-1}\sum_{n=1}^{k}\psi_{j,n}\otimes \left(\mathbfcal{I}-\mathbfcal{V}\right)^{-1}\pi_{n}\left(y_{j}\right)\\&=\sum_{j=1}^{L-1}\mathbfcal{A}_{j}y_{j},
\end{align*}

\noindent which shows that the last component of any functional $L$-lagged vector in $\mathbb{L}$ is a linear combination of the last $L-1$ components.

\end{proof}

\begin{proof}[Proof of Prop.]\ref{prop:vforecastpi}

To show that $\pmb{\Pi}$ is an orthogonal projection onto $\mathbb{L}^{\nabla}$, we need to show that it is idempotent and self-adjoint. We begin by showing the idempotent property of $\pmb{\Pi}$. First notice that $\left(\left(\mathbfcal{P}^{\nabla}\right)^{*}\mathbfcal{P}^{\nabla}\right)^{-1}$ is an $k \times k$ real-valued matrix. Now let $\pmb{v} \in \mathbb{H}^{L-1}$ and we find that

\begin{align*}
\pmb{\Pi}\left(\pmb{\Pi}\left(\pmb{v}\right)\right)&=\mathbfcal{P}^{\nabla}\left(\left(\mathbfcal{P}^{\nabla}\right)^{*}\mathbfcal{P}^{\nabla}\right)^{-1}\left(\mathbfcal{P}^{\nabla}\right)^{*}\mathbfcal{P}^{\nabla}\left(\left(\mathbfcal{P}^{\nabla}\right)^{*}\mathbfcal{P}^{\nabla}\right)^{-1}\left(\mathbfcal{P}^{\nabla}\right)^{*}\left(\pmb{v}\right)\\&=\mathbfcal{P}^{\nabla}\left(\left(\mathbfcal{P}^{\nabla}\right)^{*}\mathbfcal{P}^{\nabla}\right)^{-1}\left(\mathbfcal{P}^{\nabla}\right)^{*}\left(\pmb{v}\right)=\pmb{\Pi}\left(\pmb{v}\right)
\end{align*}

\noindent therefore $\pmb{\Pi}$ is idempotent and is a projector onto $\mathbb{L}^{\nabla}$.

Now we need to prove the orthogonality of $\pmb{\Pi}$, by showing that the operator is self-adjoint. We first determine the form of $\left(\left(\mathbfcal{P}^{\nabla}\right)^{*}\mathbfcal{P}^{\nabla}\right)^{-1}$. Notice that $\mathbf{I}_{k}=\left(\mathbfcal{P}^{\nabla}\right)^{*}\mathbfcal{P}^{\nabla}+\mathbf{B}$ where $\mathbf{I}_{k}$ is the $k \times k$ identity matrix and $\mathbf{B}=\left[\innp{\pi_{i}}{\pi_{j}}_{\mathbb{H}}\right]_{i,j=1}^{k}$. To this end, $\left(\left(\mathbfcal{P}^{\nabla}\right)^{*}\mathbfcal{P}^{\nabla}\right)^{-1}=\left(\mathbf{I}_{k}-\mathbf{B}\right)^{-1}$. We define $\mathbf{C}=\left[\innp{\pi_{i}}{\left(\mathbfcal{I}-\mathbfcal{V}\right)^{-1}\left(\pi_{j}\right)}_{\mathbb{H}}\right]_{i,j=1}^{k}$ and now multiply $\left(\mathbfcal{P}^{\nabla}\right)^{*}\mathbfcal{P}^{\nabla}\left(\mathbf{I}_{k}+\mathbf{C}\right)=\left(\mathbf{I}_{k}-\mathbf{B}\right)\left(\mathbf{I}_{k}+\mathbf{C}\right)=\mathbf{I}_{k}+\mathbf{C}-\mathbf{B}\mathbf{C}-\mathbf{B}$. Now notice that

\begin{align*}
\left[\mathbf{C}-\mathbf{B}\mathbf{C}\right]_{i,j}&=\innp{\pi_{i}}{\left(\mathbfcal{I}-\mathbfcal{V}\right)^{-1}\left(\pi_{j}\right)}_{\mathbb{H}}-\sum_{n=1}^{k}\innp{\pi_{i}}{\pi_{n}}_{\mathbb{H}}\innp{\pi_{n}}{\left(\mathbfcal{I}-\mathbfcal{V}\right)^{-1}\left(\pi_{j}\right)}_{\mathbb{H}}\\&=\innp{\pi_{i}}{\left(\mathbfcal{I}-\mathbfcal{V}\right)^{-1}\left(\pi_{j}\right)}_{\mathbb{H}}-\innp{\pi_{i}}{\sum_{n=1}^{k}\innp{\left(\mathbfcal{I}-\mathbfcal{V}\right)^{-1}\left(\pi_{j}\right)}{\pi_{n}}_{\mathbb{H}}\pi_{n}}_{\mathbb{H}}
\end{align*}
\noindent Since $\mathbb{E} \cap \mathbb{L} = \emptyset$, we have that $\left(\mathbfcal{I}-\mathbfcal{V}\right)^{-1}=\sum_{l=0}^{\infty}\mathbfcal{V}^{l}$ is a linear operator and this allows us to state that
\begin{align*}
&\innp{\pi_{i}}{\left(\mathbfcal{I}-\mathbfcal{V}\right)^{-1}\left(\pi_{j}\right)}_{\mathbb{H}}-\innp{\pi_{i}}{\sum_{n=1}^{k}\innp{\left(\mathbfcal{I}-\mathbfcal{V}\right)^{-1}\left(\pi_{j}\right)}{\pi_{n}}_{\mathbb{H}}\pi_{n}}_{\mathbb{H}}\\&=\innp{\pi_{i}}{\left(\mathbfcal{I}-\mathbfcal{V}\right)^{-1}\left(\pi_{j}\right)}_{\mathbb{H}}-\innp{\pi_{i}}{\left(\mathbfcal{I}-\mathbfcal{V}\right)^{-1}\left(\sum_{n=1}^{k}\innp{\pi_{j}}{\pi_{n}}_{\mathbb{H}}\pi_{n}\right)}_{\mathbb{H}}\\&=\innp{\pi_{i}}{\left(\mathbfcal{I}-\mathbfcal{V}\right)^{-1}\left(\pi_{j}\right)}_{\mathbb{H}}-\innp{\pi_{i}}{\left(\mathbfcal{I}-\mathbfcal{V}\right)^{-1}\left(\mathbfcal{V}\left(\pi_{j}\right)\right)}_{\mathbb{H}}\\&=\innp{\pi_{i}}{\left(\mathbfcal{I}-\mathbfcal{V}\right)^{-1}\left(\pi_{j}\right)-\left(\mathbfcal{I}-\mathbfcal{V}\right)^{-1}\left(\mathbfcal{V}\left(\pi_{j}\right)\right)}_{\mathbb{H}}\\&=\innp{\pi_{i}}{\left(\mathbfcal{I}-\mathbfcal{V}\right)^{-1}\left(\mathbfcal{I}-\mathbfcal{V}\right)\left(\pi_{j}\right)}_{\mathbb{H}}\\&=\innp{\pi_{i}}{\pi_{j}}_{\mathbb{H}}=\mathbf{B}_{i,j}
\end{align*}

\noindent and thus, we have $\mathbf{I}_{k}+\mathbf{C}-\mathbf{B}\mathbf{C}-\mathbf{B}=\mathbf{I}_{k}+\mathbf{B}-\mathbf{B}=\mathbf{I}_{k}$ which shows that $\left(\left(\mathbfcal{P}^{\nabla}\right)^{*}\mathbfcal{P}^{\nabla}\right)^{-1}=\left(\mathbf{I}_{k}+\mathbf{C}\right)$. Notice that this also implies that $\mathbf{C}$ is symmetric and we will use this fact in the following. Now let $\pmb{u} \in \mathbb{H}^{L-1}$ and notice that

\begin{align*}
\innp{\pmb{\Pi}\left(\pmb{v}\right)}{\pmb{u}}_{\mathbb{H}^{L-1}}&=\innp{\mathbfcal{P}^{\nabla}\left(\left(\mathbfcal{P}^{\nabla}\right)^{*}\mathbfcal{P}^{\nabla}\right)^{-1}\left(\mathbfcal{P}^{\nabla}\right)^{*}\left(\pmb{v}\right)}{\pmb{u}}_{\mathbb{H}^{L-1}}\\&=\innp{\mathbfcal{P}^{\nabla}\begin{bmatrix}\innp{\pmb{v}}{\pmb{\psi}_{1}^{\nabla}}_{\mathbb{H}^{L-1}}+\sum_{j=1}^{k}\innp{\pi_{1}}{\left(\mathbfcal{I}-\mathbfcal{V}\right)^{-1}\left(\pi_{j}\right)}_{\mathbb{H}}\innp{\pmb{v}}{\pmb{\psi}_{j}^{\nabla}}_{\mathbb{H}^{L-1}} \\ \vdots \\ \innp{\pmb{v}}{\pmb{\psi}_{k}^{\nabla}}_{\mathbb{H}^{L-1}}+\sum_{j=1}^{k}\innp{\pi_{k}}{\left(\mathbfcal{I}-\mathbfcal{V}\right)^{-1}\left(\pi_{j}\right)}_{\mathbb{H}}\innp{\pmb{v}}{\pmb{\psi}_{j}^{\nabla}}_{\mathbb{H}^{L-1}} \end{bmatrix}}{\pmb{u}}_{\mathbb{H}^{L-1}}\\&= \sum_{i=1}^{k}\left(\innp{\pmb{v}}{\pmb{\psi}_{i}^{\nabla}}_{\mathbb{H}^{L-1}}+ \sum_{j=1}^{k}\innp{\pi_{i}}{\left(\mathbfcal{I}-\mathbfcal{V}\right)^{-1}\left(\pi_{j}\right)}_{\mathbb{H}}\innp{\pmb{v}}{\pmb{\psi}_{j}^{\nabla}}_{\mathbb{H}^{L-1}}\right)\innp{\pmb{\psi}_{i}^{\nabla}}{\pmb{u}}_{\mathbb{H}^{L-1}}.
\end{align*}

\noindent Since $\mathbf{C}$ is symmetric we may state that $\innp{\pi_{i}}{\left(\mathbfcal{I}-\mathbfcal{V}\right)^{-1}\left(\pi_{j}\right)}_{\mathbb{H}}=\innp{\pi_{j}}{\left(\mathbfcal{I}-\mathbfcal{V}\right)^{-1}\left(\pi_{i}\right)}_{\mathbb{H}}$. As a result, we have that

\begin{align*}
&\sum_{i=1}^{k}\left(\innp{\pmb{v}}{\pmb{\psi}_{i}^{\nabla}}_{\mathbb{H}^{L-1}}+ \sum_{j=1}^{k}\innp{\pi_{i}}{\left(\mathbfcal{I}-\mathbfcal{V}\right)^{-1}\left(\pi_{j}\right)}_{\mathbb{H}}\innp{\pmb{v}}{\pmb{\psi}_{j}^{\nabla}}_{\mathbb{H}^{L-1}}\right)\innp{\pmb{\psi}_{i}^{\nabla}}{\pmb{u}}_{\mathbb{H}^{L-1}}\\&=\sum_{j=1}^{k}\left(\innp{\pmb{u}}{\pmb{\psi}_{j}^{\nabla}}_{\mathbb{H}^{L-1}}+ \sum_{i=1}^{k}\innp{\pi_{j}}{\left(\mathbfcal{I}-\mathbfcal{V}\right)^{-1}\left(\pi_{i}\right)}_{\mathbb{H}}\innp{\pmb{u}}{\pmb{\psi}_{i}^{\nabla}}_{\mathbb{H}^{L-1}}\right)\innp{\pmb{\psi}_{j}^{\nabla}}{\pmb{v}}_{\mathbb{H}^{L-1}}\\&=\innp{\pmb{v}}{\mathbfcal{P}^{\nabla}\begin{bmatrix}\innp{\pmb{u}}{\pmb{\psi}_{1}^{\nabla}}_{\mathbb{H}^{L-1}}+ \sum_{i=1}^{k}\innp{\pi_{1}}{\left(\mathbfcal{I}-\mathbfcal{V}\right)^{-1}\left(\pi_{i}\right)}_{\mathbb{H}}\innp{\pmb{u}}{\pmb{\psi}_{i}^{\nabla}}_{\mathbb{H}^{L-1}} \\ \vdots \\ \innp{\pmb{u}}{\pmb{\psi}_{k}^{\nabla}}_{\mathbb{H}^{L-1}}+ \sum_{i=1}^{k}\innp{\pi_{k}}{\left(\mathbfcal{I}-\mathbfcal{V}\right)^{-1}\left(\pi_{i}\right)}_{\mathbb{H}}\innp{\pmb{u}}{\pmb{\psi}_{i}^{\nabla}}_{\mathbb{H}^{L-1}} \end{bmatrix}}_{\mathbb{H}^{L-1}}\\&=\innp{\pmb{v}}{\pmb{\Pi}\left(\pmb{u}\right)}_{\mathbb{H}^{L-1}}.
\end{align*}

\noindent Thus $\pmb{\Pi}$ is self-adjoint and an orthogonal projection onto $\mathbb{L}^{\nabla}$. 

\end{proof}

\begin{proof}[Proof of Prop.]\ref{prop:vforecastlast}

Recall that for some $\pmb{x} \in \mathbb{L}$ we have $\pmb{y}^{\nabla}=\pmb{\Pi}\pmb{x}^{\Delta}$. Now we wish to show that the last component of $\pmb{y}$, defined as $y_{L}$, has the expression $y_{L}=\sum_{j=1}^{L-1}\mathbfcal{A}_{j}x_{j}^{\Delta}$. Using Theorem \ref{thm:rforecast} and the definition of $\pmb{\Pi}$, notice that

\begin{align*}
y_{L}=\sum_{j=1}^{L-1}\mathbfcal{A}_{j}y_{j}^{\nabla}=\sum_{j=1}^{L-1}\mathbfcal{A}_{j}\left(\mathbfcal{P}^{\nabla}\left(\mathbf{I}_{k}+\mathbf{C}\right)\begin{bmatrix}\innp{\pmb{x}^{\Delta}}{\pmb{\psi}_{1}^{\nabla}}_{\mathbb{H}^{L-1}} \\ \vdots \\ \innp{\pmb{x}^{\Delta}}{\pmb{\psi}_{k}^{\nabla}}_{\mathbb{H}^{L-1}} \end{bmatrix}\right)_{j}=u+v
\end{align*}

\noindent where

\begin{align*}
u&=\sum_{j=1}^{L-1}\mathbfcal{A}_{j}\left(\sum_{n=1}^{k}\innp{\pmb{x}^{\Delta}}{\pmb{\psi}_{n}^{\nabla}}_{\mathbb{H}^{L-1}}\psi_{j,n}\right)\\ v&=\sum_{j=1}^{L-1}\mathbfcal{A}_{j}\left(\mathbfcal{P}^{\nabla}\begin{bmatrix}\sum_{n=1}^{k}\innp{\pi_{1}}{\left(\mathbfcal{I}-\mathbfcal{V}\right)^{-1}\left(\pi_{n}\right)}_{\mathbb{H}}\innp{\pmb{x}^{\Delta}}{\pmb{\psi}_{n}^{\nabla}}_{\mathbb{H}^{L-1}}\\ \vdots \\\sum_{n=1}^{k}\innp{\pi_{k}}{\left(\mathbfcal{I}-\mathbfcal{V}\right)^{-1}\left(\pi_{n}\right)}_{\mathbb{H}}\innp{\pmb{x}^{\Delta}}{\pmb{\psi}_{n}^{\nabla}}_{\mathbb{H}^{L-1}}  \end{bmatrix}\right)_{j}.
\end{align*}

\noindent Now notice that

\begin{align*}
u&=\sum_{j=1}^{L-1}\sum_{n=1}^{k}\psi_{j,n} \otimes \left(\mathbfcal{I}-\mathbfcal{V}\right)^{-1}\left(\pi_{n}\right)\left(\sum_{i=1}^{k}\innp{\pmb{x}^{\Delta}}{\pmb{\psi}_{i}^{\nabla}}_{\mathbb{H}^{L-1}}\psi_{j,i}\right)\\&=\sum_{n=1}^{k}\sum_{i=1}^{d}\innp{\pmb{x}^{\Delta}}{\pmb{\psi}_{i}^{\nabla}}_{\mathbb{H}^{L-1}}\innp{\pmb{\psi}_{i}^{\nabla}}{\pmb{\psi}_{n}^{\nabla}}_{\mathbb{H}^{L-1}}\left(\mathbfcal{I}-\mathbfcal{V}\right)^{-1}\left(\pi_{n}\right)\\ &=\sum_{n=1}^{k}\sum_{i=1}^{k}\innp{\pmb{x}^{\Delta}}{\pmb{\psi}_{i}^{\nabla}}_{\mathbb{H}^{L-1}}\left(\mathbf{I}_{i,n}-\innp{\pi_{i}}{\pi_{n}}_{\mathbb{H}}\right)\left(\mathbfcal{I}-\mathbfcal{V}\right)^{-1}\left(\pi_{n}\right)\\ &= \sum_{n=1}^{k}\innp{\pmb{x}^{\Delta}}{\pmb{\psi}_{n}^{\nabla}}_{\mathbb{H}^{L-1}}\left(\mathbfcal{I}-\mathbfcal{V}\right)^{-1}\left(\pi_{n}\right)-\sum_{i=1}^{k}\sum_{l_{1}=1}^{L-1}\innp{x^{\Delta}_{l_{1}}}{\psi_{l_{1},i}}_{\mathbb{H}}\left(\mathbfcal{I}-\mathbfcal{V}\right)^{-1}\mathbfcal{V}\left(\pi_{i}\right)\\ &= \sum_{l_{2}=1}^{L-1}\mathbfcal{A}_{l_{2}}x^{\Delta}_{l_{2}}-\mathbfcal{V}\left(\sum_{l_{1}=1}^{L-1}\mathbfcal{A}_{l_{1}}x^{\Delta}_{l_{1}}\right).
\end{align*}

\noindent In addition, notice that
\begin{align*}
v&=\sum_{j=1}^{L-1}\mathbfcal{A}_{j}\left(\sum_{n_{1}=1}^{k}\sum_{n_{2}=1}^{k}\innp{\pi_{n_{1}}}{\left(\mathbfcal{I}-\mathbfcal{V}\right)^{-1}\left(\pi_{n_{2}}\right)}_{\mathbb{H}}\innp{\pmb{x}^{\Delta}}{\pmb{\psi}_{n_{2}}^{\nabla}}_{\mathbb{H}^{L-1}}\pmb{\psi}_{n_{1}}^{\nabla}\right)_{j}\\&=\sum_{n_{3}=1}^{k}\sum_{n_{1}=1}^{k}\sum_{n_{2}=1}^{k}\innp{\pi_{n_{1}}}{\left(\mathbfcal{I}-\mathbfcal{V}\right)^{-1}\left(\pi_{n_{2}}\right)}_{\mathbb{H}}\innp{\pmb{x}^{\Delta}}{\pmb{\psi}_{n_{2}}^{\nabla}}_{\mathbb{H}^{L-1}}\innp{\pmb{\psi}_{n_{1}}^{\nabla}}{\pmb{\psi}_{n_{3}}^{\nabla}}_{\mathbb{H}^{L-1}}\left(\mathbfcal{I}-\mathbfcal{V}\right)^{-1}\left(\pi_{n_{3}}\right)\\&=\sum_{n_{3}=1}^{k}\sum_{n_{1}=1}^{k}\sum_{n_{2}=1}^{k}\innp{\pi_{n_{1}}}{\left(\mathbfcal{I}-\mathbfcal{V}\right)^{-1}\left(\pi_{n_{2}}\right)}_{\mathbb{H}}\innp{\pmb{x}^{\Delta}}{\pmb{\psi}_{n_{2}}^{\nabla}}_{\mathbb{H}^{L-1}}\left(\mathbf{I}_{n_{1},n_{3}}-\innp{\pi_{n_{1}}}{\pi_{n_{3}}}_{\mathbb{H}}\right)\left(\mathbfcal{I}-\mathbfcal{V}\right)^{-1}\left(\pi_{n_{3}}\right)\\&=\sum_{n_{1}=1}^{k}\sum_{n_{2}=1}^{k}\innp{\pmb{x}^{\Delta}}{\pmb{\psi}_{n_{2}}^{\nabla}}_{\mathbb{H}^{L-1}}\innp{\pi_{n_{1}}}{\left(\mathbfcal{I}-\mathbfcal{V}\right)^{-1}\left(\pi_{n_{2}}\right)}_{\mathbb{H}}\left(\left(\mathbfcal{I}-\mathbfcal{V}\right)^{-1}\left(\pi_{n_{1}}\right)-\left(\mathbfcal{I}-\mathbfcal{V}\right)^{-1}\mathbfcal{V}\left(\pi_{n_{1}}\right)\right)\\&=\sum_{n_{1}=1}^{k}\sum_{n_{2}=1}^{k}\innp{\pmb{x}^{\Delta}}{\pmb{\psi}_{n_{2}}^{\nabla}}_{\mathbb{H}^{L-1}}\innp{\pi_{n_{1}}}{\left(\mathbfcal{I}-\mathbfcal{V}\right)^{-1}\left(\pi_{n_{2}}\right)}_{\mathbb{H}}\left(\left(\mathbfcal{I}-\mathbfcal{V}\right)^{-1}\left(\mathbfcal{I}-\mathbfcal{V}\right)\left(\pi_{n_{1}}\right)\right)\\&=\mathbfcal{V}\left(\sum_{n_{2}=1}^{k}\innp{\pmb{x}^{\Delta}}{\pmb{\psi}_{n_{2}}^{\nabla}}_{\mathbb{H}^{L-1}}\left(\mathbfcal{I}-\mathbfcal{V}\right)^{-1}\left(\pi_{n_{2}}\right)\right)\\ &= \mathbfcal{V}\left(\sum_{j=1}^{L-1}\mathbfcal{A}_{j}x^{\Delta}_{j}\right).
\end{align*}

\noindent As a result, we find that

\begin{equation*}
y_{L}=\sum_{j=1}^{L-1}\mathbfcal{A}_{j}y^{\nabla}_{j}=u+v=\sum_{j=1}^{L-1}\mathbfcal{A}_{j}x_{j}^{\Delta}-\mathbfcal{V}\left(\sum_{j=1}^{L-1}\mathbfcal{A}_{j}x^{\Delta}_{j}\right)+\mathbfcal{V}\left(\sum_{j=1}^{L-1}\mathbfcal{A}_{j}x^{\Delta}_{j}\right)=\sum_{j=1}^{L-1}\mathbfcal{A}_{j}x^{\Delta}_{j}.
\end{equation*}

\noindent Thus we see that $y_{L}=\sum_{j=1}^{L-1}\mathbfcal{A}_{j}y^{\nabla}_{j}=\sum_{j=1}^{L-1}\mathbfcal{A}_{j}x^{\Delta}_{j}$.

\end{proof}

\begin{proof}[Proof of Lemma]\ref{lem:opimp}

First, we notice that

\begin{equation*}
\innp{\left(\mathbfcal{I}-\mathbfcal{V}\right)^{-1}\left(\pi_{n}\right)}{\nu_{i}}_{\mathbb{H}}=\sum_{l=0}^{\infty}\innp{\mathbfcal{V}^{l}\left(\pi_{n}\right)}{\nu_{i}}_{\mathbb{H}}.
\end{equation*}

\noindent As such, we use induction to show the desired result of $\left[\innp{\mathbfcal{V}^{l}\left(\pi_{n}\right)}{\nu_{i}}_{\mathbb{H}}\right]_{i=1,\dots,d}^{n=1,\dots,k}=\mathbf{G}^{\frac{1}{2}}\mathbf{D}\left(\mathbf{D}^{\top}\mathbf{D}\right)^{l}$. For our base case, let $l=1$, then

\begin{align*}
\innp{\mathbfcal{V}\left(\pi_{n}\right)}{\nu_{i}}_{\mathbb{H}}=\sum_{n_{1}=1}^{k}\innp{\pi_{n}}{\pi_{n_{1}}}_{\mathbb{H}}\innp{\pi_{n_{1}}}{\nu_{i}}_{\mathbb{H}}=\left(\mathbf{G}^{\frac{1}{2}}\mathbf{D}\mathbf{D}^{\top}\mathbf{D}\right)_{i,n}.
\end{align*}

\noindent From here, we state our inductive assumption that $\left[\innp{\mathbfcal{V}^{l}\left(\pi_{n}\right)}{\nu_{i}}_{\mathbb{H}}\right]_{i=1,\dots,d}^{n=1,\dots,k}=\mathbf{G}^{\frac{1}{2}}\mathbf{D}\left(\mathbf{D}^{\top}\mathbf{D}\right)^{l}$. Now notice that

\begin{align*}
\innp{\mathbfcal{V}^{l+1}\left(\pi_{n}\right)}{\nu_{i}}_{\mathbb{H}}&=\sum_{n_{1}=1}^{k}\cdots \sum_{n_{l+1}=1}^{k}\innp{\pi_{n}}{\pi_{n_{l+1}}}_{\mathbb{H}}\innp{\pi_{n_{l+1}}}{\pi_{n_{l}}}_{\mathbb{H}}\cdots \innp{\pi_{n_{2}}}{\pi_{n_{1}}}_{\mathbb{H}}\innp{\pi_{n_{1}}}{\nu_{i}}_{\mathbb{H}}\\&=\left(\mathbf{G}^{\frac{1}{2}}\mathbf{D}\left(\mathbf{D}^{\top}\mathbf{D}\right)^{l+1}\right)_{i,n}=\left(\mathbf{G}^{\frac{1}{2}}\mathbf{D}\mathbf{D}^{\top}\mathbf{D}\left(\mathbf{D}^{\top}\mathbf{D}\right)^{l}\right)_{i,n}=\innp{\mathbfcal{V}\left(\mathbfcal{V}^{l}\left(\pi_{n}\right)\right)}{\nu_{i}}_{\mathbb{H}}.
\end{align*}

\noindent which shows the desired result. As such, we see that

\begin{equation*}
\innp{\left(\mathbfcal{I}-\mathbfcal{V}\right)^{-1}\left(\pi_{n}\right)}{\nu_{i}}_{\mathbb{H}}=\sum_{l=0}^{\infty}\innp{\mathbfcal{V}^{l}\left(\pi_{n}\right)}{\nu_{i}}_{\mathbb{H}}=\mathbf{G}^{\frac{1}{2}}\mathbf{D}\left(\sum_{l=0}^{\infty}\left(\mathbf{D}^{\top}\mathbf{D}\right)^{l}\right)_{i,n}
\end{equation*}

\end{proof}

\begin{proof}[Proof of Thm.]\ref{thm:rforecastimp}

Let $f \in \mathbb{H}$ and recall that

\begin{equation*}
\mathbfcal{A}_{j}(f)=\sum_{n=1}^{k}\innp{\psi_{j,n}}{f}_{\mathbb{H}}\left(\mathbfcal{I}-\mathbfcal{V}\right)^{-1}\left(\pi_{n}\right).
\end{equation*}

\noindent Now using the result of Lemma \ref{lem:opimp}

\begin{align*}
\left(\mathbf{G}^{\frac{1}{2}}\mathbf{A}_{j}\mathbf{c}_{f}\right)_{i}&=\left(\mathbf{G}^{\frac{1}{2}}\mathbf{D}\left(\sum_{l=0}^{\infty}\left(\mathbf{D}^{\top}\mathbf{D}\right)^{l}\right)\mathbf{E}_{j}^{\top}\mathbf{G}^{\frac{1}{2}}\mathbf{c}_{f}\right)_{i}\\&=\innp{\sum_{n=1}^{k}\innp{\psi_{j,n}}{f}_{\mathbb{H}}\left(\mathbfcal{I}-\mathbfcal{V}\right)^{-1}\left(\pi_{n}\right)}{\nu_{i}}_{\mathbb{H}}\\&=\innp{\mathbfcal{A}_{j}\left(f\right)}{\nu_{i}}_{\mathbb{H}}.
\end{align*}




\noindent This implies that the matrices that implement the R-forecasting algorithm have the form of

\begin{equation*}
\mathbf{A}_{j}=\mathbf{D}\left(\sum_{l=0}^{\infty}\left(\mathbf{D}^{\top}\mathbf{D}\right)^{l}\right)\mathbf{E}_{j}^{\top}\mathbf{G}^{\frac{1}{2}},\quad j=1,\dots,L-1.
\end{equation*}



\end{proof}

\begin{proof}[Proof of Thm.]\ref{thm:vforecastimp}




Notice that

\begin{align*}
&\innp{\pmb{\Pi}\left(\pmb{\phi}_{j}^{\nabla}\right)}{\pmb{\phi}_{i}^{\nabla}}_{\mathbb{H}^{L-1}}=\innp{\mathbfcal{P}^{\nabla}\left(\left(\mathbf{I}_{k}+\mathbf{C}\right)\begin{bmatrix}\innp{\pmb{\phi}_{j}^{\nabla}}{\pmb{\psi}_{1}^{\nabla}}_{\mathbb{H}^{L-1}}\\ \vdots \\ \innp{\pmb{\phi}_{j}^{\nabla}}{\pmb{\psi}_{k}^{\nabla}}_{\mathbb{H}^{L-1}} \end{bmatrix}\right)}{\pmb{\phi}_{i}^{\nabla}}_{\mathbb{H}^{L-1}}\\&=\innp{\mathbfcal{P}^{\nabla}\left(\begin{bmatrix}\innp{\pmb{\phi}_{j}^{\nabla}}{\pmb{\psi}_{1}^{\nabla}}_{\mathbb{H}^{L-1}}\\ \vdots \\ \innp{\pmb{\phi}_{j}^{\nabla}}{\pmb{\psi}_{k}^{\nabla}}_{\mathbb{H}^{L-1}} \end{bmatrix}\right)+\mathbfcal{P}^{\nabla}\left(\begin{bmatrix}\sum_{n=1}^{k}\innp{\pi_{1}}{\left(\mathbfcal{I}-\mathbfcal{V}\right)^{-1}\left(\pi_{n}\right)}_{\mathbb{H}^{L-1}}\innp{\pmb{\phi}_{j}^{\nabla}}{\pmb{\psi}_{n}^{\nabla}}_{\mathbb{H}^{L-1}}\\ \vdots \\ \sum_{n=1}^{k}\innp{\pi_{k}}{\left(\mathbfcal{I}-\mathbfcal{V}\right)^{-1}\left(\pi_{n}\right)}_{\mathbb{H}^{L-1}}\innp{\pmb{\phi}_{j}^{\nabla}}{\pmb{\psi}_{n}^{\nabla}}_{\mathbb{H}^{L-1}} \end{bmatrix}\right)}{\pmb{\phi}_{i}^{\nabla}}_{\mathbb{H}^{L-1}}\\&=\sum_{l=1}^{k}\innp{\pmb{\phi}_{j}^{\nabla}}{\pmb{\psi}_{l}^{\nabla}}_{\mathbb{H}^{L-1}}\innp{\pmb{\psi}_{l}^{\nabla}}{\pmb{\phi}_{i}^{\nabla}}_{\mathbb{H}^{L-1}}+\sum_{l=1}^{k}\sum_{n=1}^{k}\innp{\pmb{\phi}_{i}^{\nabla}}{\pmb{\psi}_{l}^{\nabla}}_{\mathbb{H}^{L-1}}\innp{\pi_{l}}{\left(\mathbfcal{I}-\mathbfcal{V}\right)^{-1}\left(\pi_{n}\right)}_{\mathbb{H}^{L-1}}\innp{\pmb{\psi}_{n}^{\nabla}}{\pmb{\phi}_{j}^{\nabla}}_{\mathbb{H}^{L-1}}\\&=\left(\left(\mathbf{H}^{\nabla}\right)^{\frac{1}{2}}\mathbf{F}\mathbf{F}^{\top}\left(\mathbf{H}^{\nabla}\right)^{\frac{1}{2}}+\left(\mathbf{H}^{\nabla}\right)^{\frac{1}{2}}\mathbf{F}\mathbf{D}^{\top}\mathbf{D}\sum_{l=0}^{\infty}\left(\mathbf{D}^{\top}\mathbf{D}\right)^{l}\mathbf{F}^{\top}\left(\mathbf{H}^{\nabla}\right)^{\frac{1}{2}}\right)_{i,j}\\&=\left(\left(\mathbf{H}^{\nabla}\right)^{\frac{1}{2}}\left(\mathbf{F}\mathbf{F}^{\top}\left(\mathbf{H}^{\nabla}\right)^{\frac{1}{2}}+\mathbf{F}\mathbf{D}^{\top}\mathbf{D}\sum_{l=0}^{\infty}\left(\mathbf{D}^{\top}\mathbf{D}\right)^{l}\mathbf{F}^{\top}\left(\mathbf{H}^{\nabla}\right)^{\frac{1}{2}}\right)\right)_{i,j}
\end{align*}

\noindent Now notice that for some $\pmb{f} \in \mathbb{H}_{d}^{L-1}$, we have

\begin{equation*}
\innp{\pmb{\Pi}\left(\pmb{f}\right)}{\pmb{\phi}_{i}^{\nabla}}_{\mathbb{H}^{L-1}}=\left(\left(\mathbf{H}^{\nabla}\right)^{\frac{1}{2}}\left(\mathbf{F}\mathbf{F}^{\top}\left(\mathbf{H}^{\nabla}\right)^{\frac{1}{2}}\mathbf{c}_{\pmb{f}}+\mathbf{F}\mathbf{D}^{\top}\mathbf{D}\sum_{l=0}^{\infty}\left(\mathbf{D}^{\top}\mathbf{D}\right)^{l}\mathbf{F}^{\top}\left(\mathbf{H}^{\nabla}\right)^{\frac{1}{2}}\mathbf{c}_{\pmb{f}}\right)\right)_{i}.
\end{equation*}

\noindent This implies that the matrix that implements $\pmb{\Pi}$ is

\begin{equation*}
\mathbf{P}=\left(\mathbf{F}\mathbf{F}^{\top}+\mathbf{F}\mathbf{D}^{\top}\mathbf{D}\sum_{l=0}^{\infty}\left(\mathbf{D}^{\top}\mathbf{D}\right)^{l}\mathbf{F}^{\top}\right)\left(\mathbf{H}^{\nabla}\right)^{\frac{1}{2}}.
\end{equation*}


\end{proof}

\begin{proof}[Proof of Thm.]\ref{thm:fullvforecastimp}

Let $\pmb{x} \in \mathbb{L}$, then using the results of Theorems \ref{prop:vforecastpi}, \ref{prop:vforecastlast}, \ref{thm:rforecastimp}, and \ref{thm:vforecastimp}, we find the coefficients of $\mathbfcal{Q}\pmb{x}$ are given by

\begin{equation*}
\mathbf{c}_{\mathbfcal{Q}}(\pmb{x})=\begin{pmatrix}\mathbf{c}_{\pmb{y}^{\nabla}} \\ \sum_{j=1}^{L-1}\mathbf{A}_{j}\mathbf{c}_{y_{j}^{\nabla}}\end{pmatrix} =\begin{pmatrix}\mathbf{P}\mathbf{c}_{\pmb{x}^{\Delta}} \\ \sum_{j=1}^{L-1}\mathbf{A}_{j}\mathbf{c}_{x_{j}^{\Delta}}\end{pmatrix}.
\end{equation*}

\end{proof}









\subsection*{Simulated Data}
In order to further clarify what varying simulation setups from the manuscript look like, we consider a subset of the different cases where we change the periodicity in the data, the trend, and the noise structure. We hold $L=40$, $N=200$, and $M=80$ constant and we generate the visuanimation of Figure \ref{fig:simani} which shows the testing set true signal we are trying to predict in plot (A), the rolling forecast prediction of the testing set using the method of \cite{hyndman2007} (B), the rolling forecast prediction using V-forecasting in plot (C), and the rolling forecast prediction using R-forecasting in plot (D). The first four frames of the animation are concerned with simulated data where only periodicity is present, the next four are concerned with cases where there is a mix of periodicity and trend, and the last four frames consider setups where there is only trend in the data.

\begin{figure}[H]
	\centering
	\animategraphics[controls,width=.7\textwidth]{.25}{animation_signal_4}{0}{11}	
	\caption{(A): testing set true signal, (B): forecast of testing set using method of \cite{hyndman2007}, (C): V-forecast of testing set, (D): R-forecast of testing set. The last four figures in that animuation were generated using \pkg{rainbow} \citep{rainbowpackage}.}
	\label{fig:simani}
\end{figure}

\noindent For the last four figures in the visuanimation, warmer colors are indicative of observations from earlier time points while cooler colors are indicative of observations of later time points.

\subsection*{Real Data}
\begin{figure}[H]
\begin{center}
    \begin{subfigure}[b]{0.4\textwidth}
	\includegraphics[page=1,width=\textwidth]{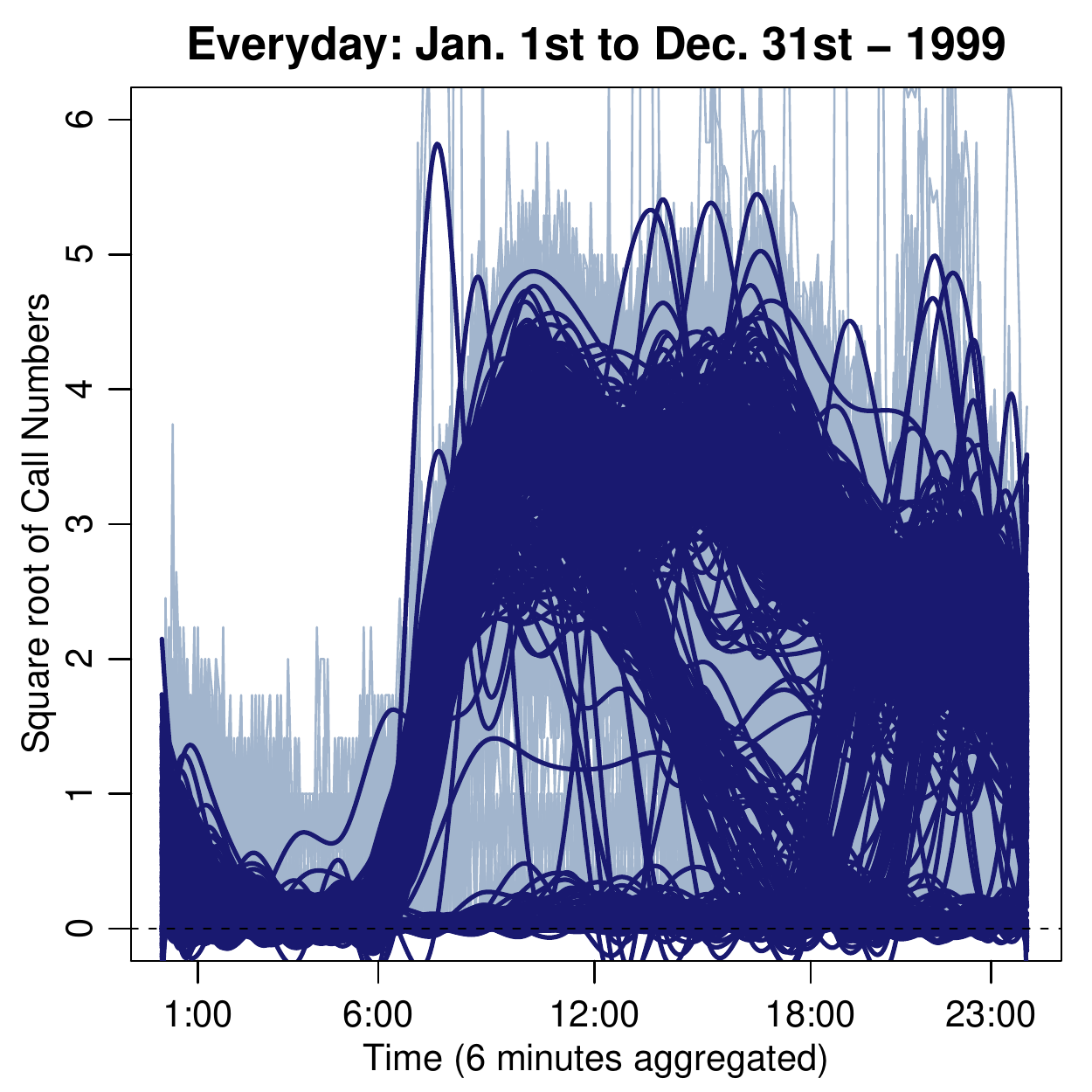}
    \end{subfigure}
    \begin{subfigure}[b]{0.4\textwidth}
	\includegraphics[page=2,width=\textwidth]{Figure1}
    \end{subfigure}
\caption{The number of calls to a call center between January 1\textit{st} to December 31\textit{st} in the year 1999.}
\label{fig:motivating_call}
\end{center}
\end{figure}

Here, we give further justification to the claims in the manuscript that there exists a strong weekly periodicity in the call center data with no trend and that there's a mix of periodicity and trend present in the NDVI data. We do not argue the lack of periodicity and presence of trend in the mortality rate data as this can be inferred from plot (C) of Figure 2 of the manuscript. This section is to be viewed as a summary of results in \cite{haghbin2019} and all of the following figures are drawn directly from that work and the corresponding supplement of \cite{haghbin2019supp}. We begin with analysis of the call center data. To start, we offer Figure \ref{fig:motivating_call} which shows a clear weekly pattern is present in the raw data where the function observed, depends on the day of the week.

\begin{figure}[H]
\begin{center}
    \begin{subfigure}[b]{0.35\textwidth}
	\includegraphics[page=1,width=\textwidth]{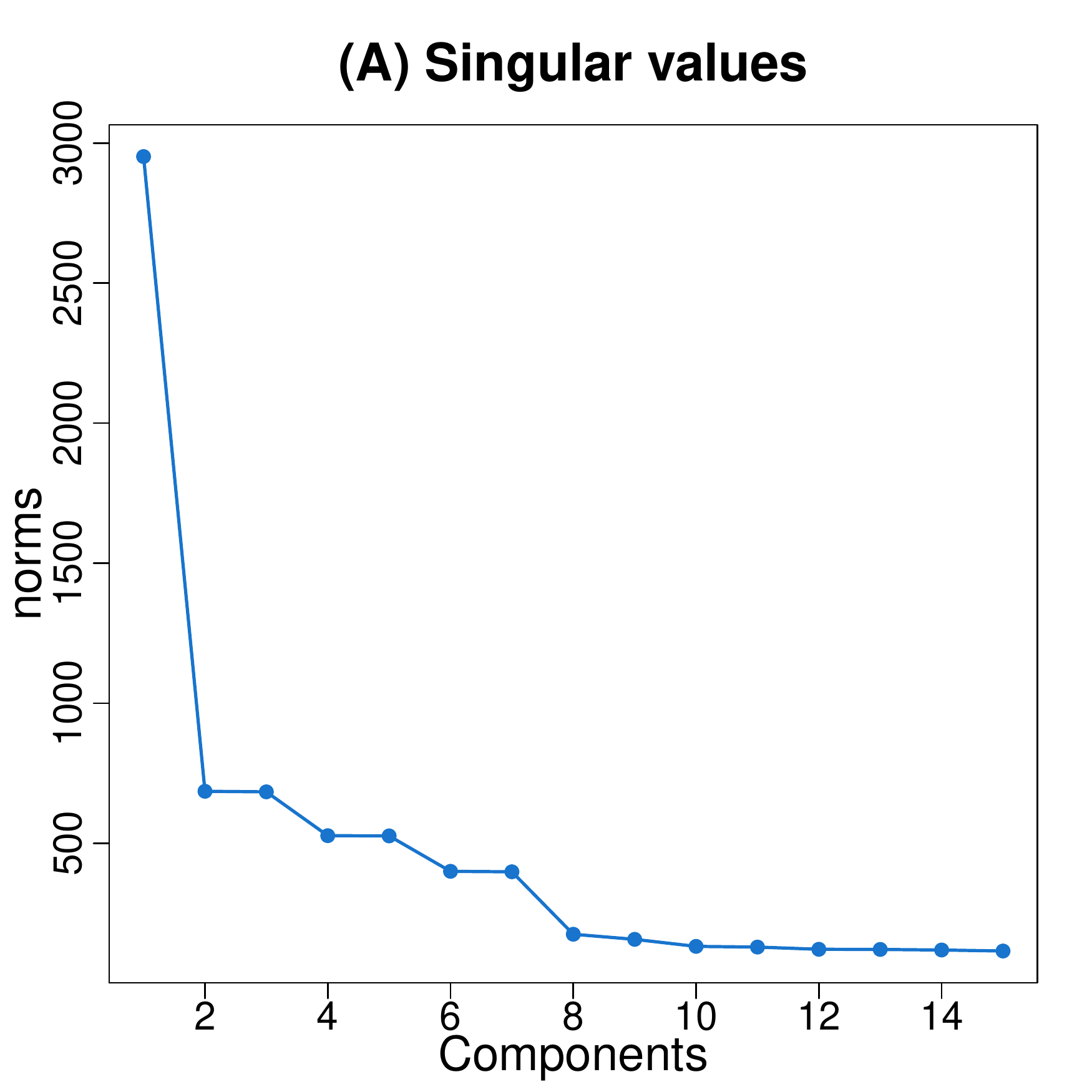}
    \end{subfigure}
    \begin{subfigure}[b]{0.35\textwidth}
	\includegraphics[page=2,width=\textwidth]{FSSA-calls}
    \end{subfigure}
        \begin{subfigure}[b]{0.35\textwidth}
	\includegraphics[page=3,width=\textwidth]{FSSA-calls}
    \end{subfigure}
        \begin{subfigure}[b]{0.35\textwidth}
	\includegraphics[page=4,width=\textwidth]{FSSA-calls}
    \end{subfigure}
\caption{(A): singular values, (B): $w$-correlation matrix, (C): right singular vectors, (D): pair plots of right singular vectors. For more information on these plots, we refer to \cite{haghbin2019}.}
\label{fig:fssacallvecs}
\end{center}
\end{figure}

Applying FSSA with a lag of $28$ to the call center data, just like in \cite{haghbin2019}, gives us the singular values of Figure \ref{fig:fssacallvecs} plot (A), the $w$-correlation matrix of plot (B), right singular vectors of plot (C), and pair plot of singular vectors (D).

\begin{figure}[H]
\begin{center}
    \begin{subfigure}[b]{0.4\textwidth}
	\includegraphics[page=5,width=\textwidth]{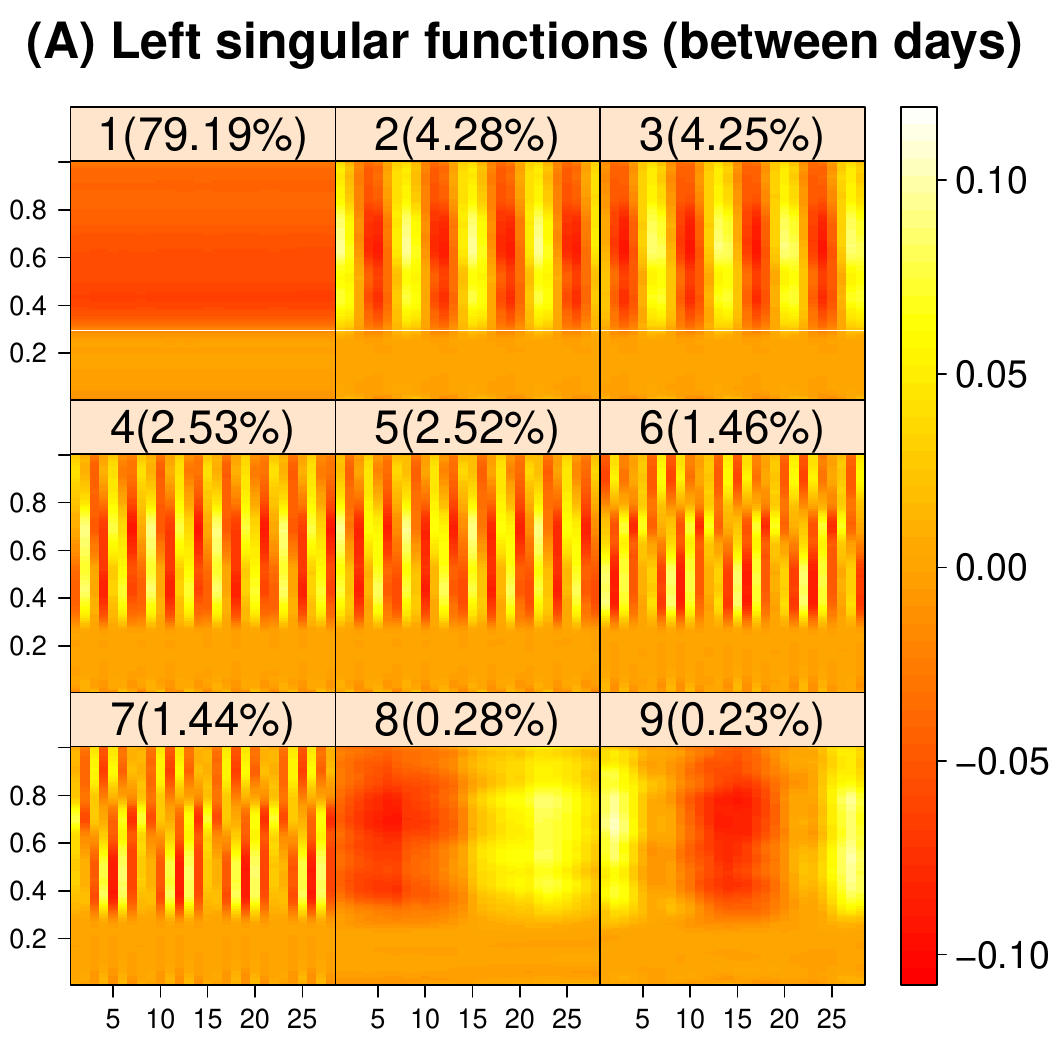}
    \end{subfigure}
    \begin{subfigure}[b]{0.4\textwidth}
	\includegraphics[page=6,width=\textwidth]{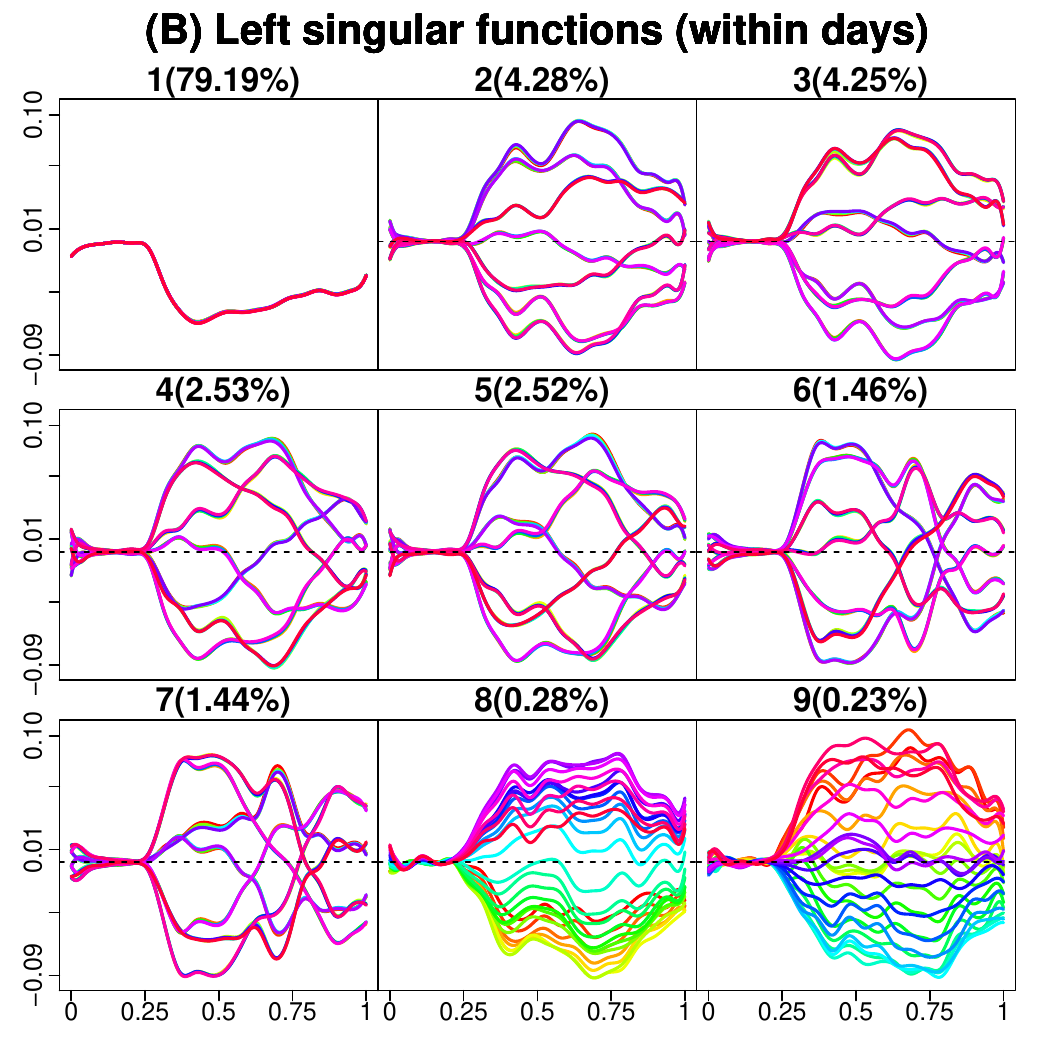}
    \end{subfigure}
\caption{(A): heat map of left singular functions, (B): left singular functions.}
\label{fig:fssacallfunc}
\end{center}
\end{figure}

\noindent We see from Figure \ref{fig:fssacallvecs}, plots (A) and (B) that there are seven components that do not correspond to noise and plots (C) and (D) show that components two through 7 reflect weekly periodicty. We see that plot (C) has oscillations in the weights that are multiplied by each left singular function indicating periodicity. Futhermore, the pair plots of 2 vs. 3, 4 vs. 5, and 6 vs. 7 in plot (D) of Figure \ref{fig:fssacallvecs} clearly shows the weekly pattern as according to the 7 sharp corners in each plot. To further argue a seven day weekly pattern, we plot the resulting left singular functions in Figure \ref{fig:fssacallfunc}.

\noindent Here, we clearly see a weekly periodicity in components two through seven where we find the oscillatory behavior in plot (A) between days and we count seven distinct curves in subplots two through seven of plot (B). From this analysis, we confirm the presence of highly periodic components in the call center data and we also notice that there are no significant trend components to be found.

\begin{figure}[H]
\begin{center}
    \begin{subfigure}[b]{0.5\textwidth}
	\includegraphics[page=1,width=\textwidth]{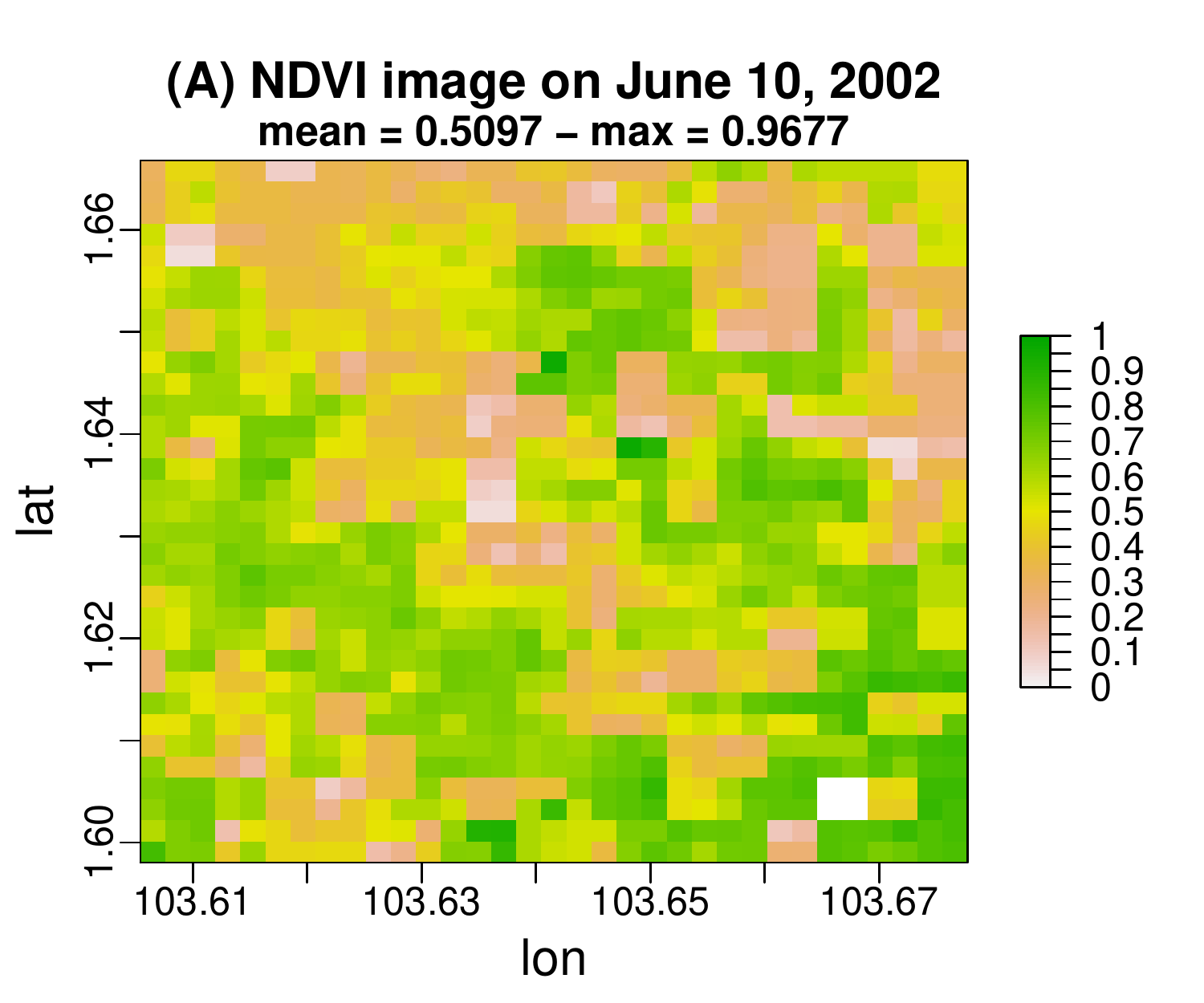}
    \end{subfigure}
    \begin{subfigure}[b]{0.49\textwidth}
	\includegraphics[page=2,width=\textwidth]{NDVI-justification}
    \end{subfigure}
\caption{(A): sample NDVI image, (B): the corresponding estimated density of NDVI}
\label{fig:ndvijust}
\end{center}
\end{figure}

We now argue the existence of periodic and trend components in the NDVI data. We first give an example of an NDVI image and the corresponding density of NDVI we estimate that are shown in Figure \ref{fig:ndvijust}.

\begin{figure}[H]
\begin{center}
	\begin{subfigure}[b]{0.33\textwidth}
		\includegraphics[width=\textwidth]{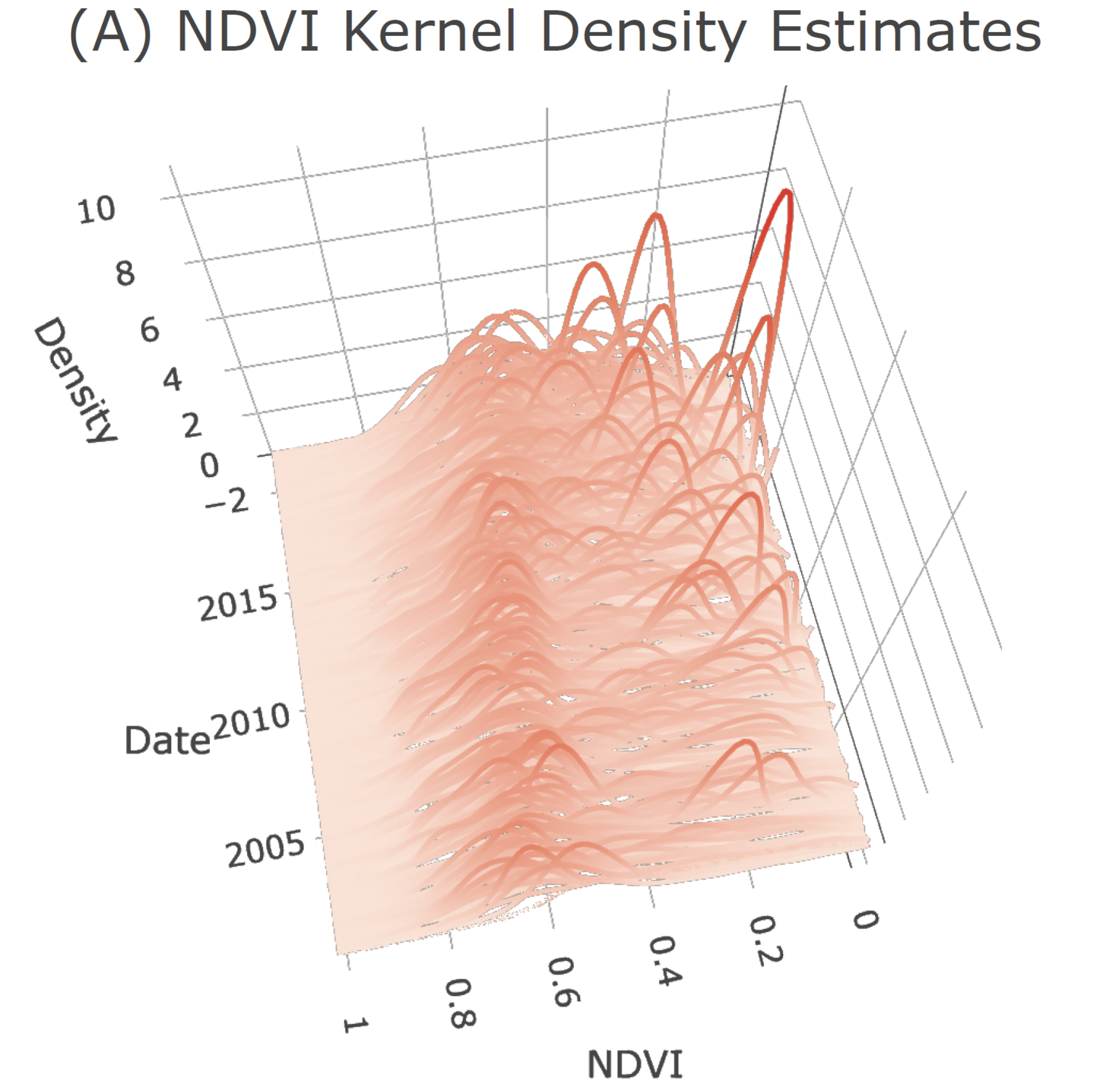}
		\label{fig:NDVI-A}
	\end{subfigure}	
	\begin{subfigure}[b]{0.33\textwidth}
		\includegraphics[page=1,width=\textwidth]{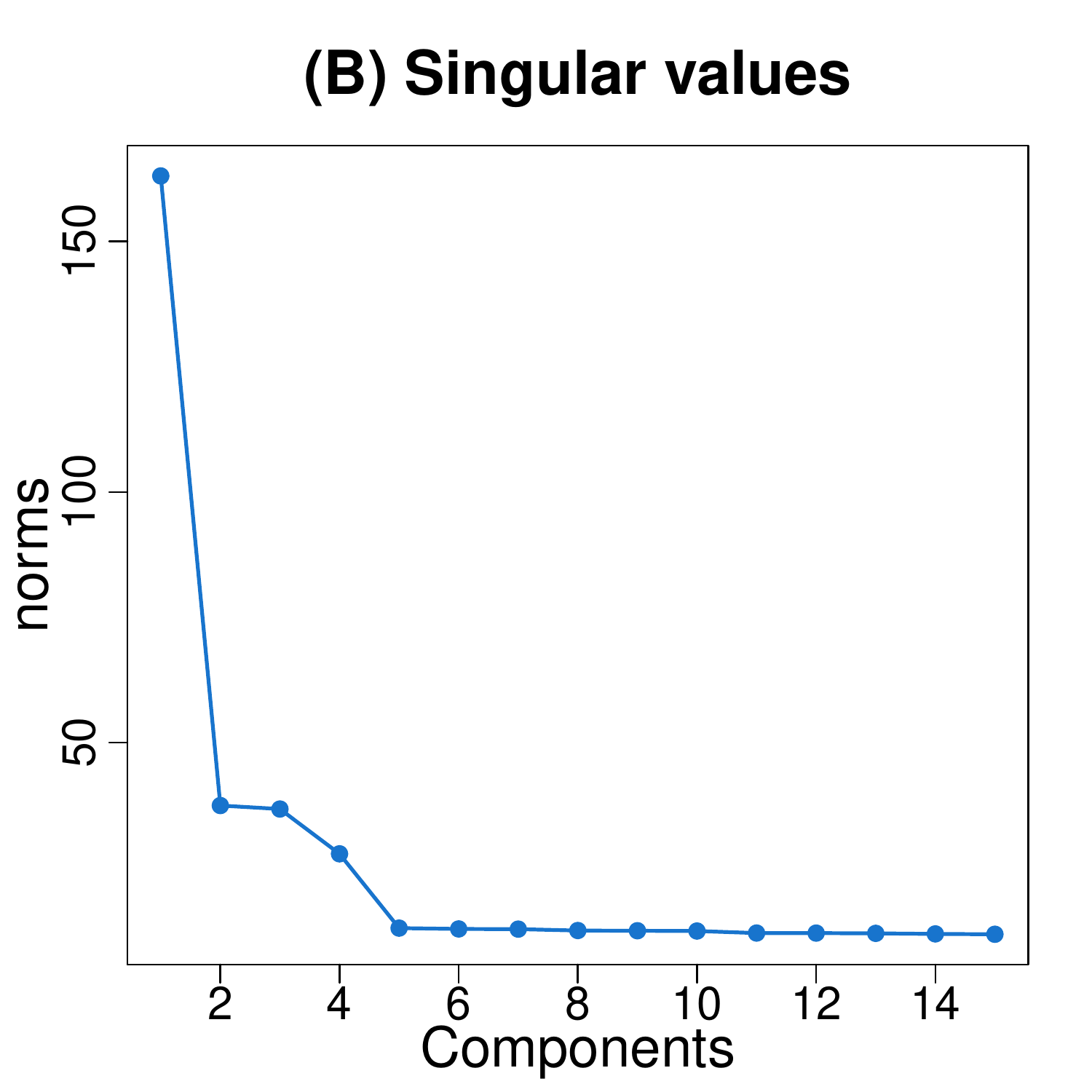}
		\label{fig:NDVI-B}
	\end{subfigure}
	\begin{subfigure}[b]{0.32\textwidth}
		\includegraphics[page=2,width=\textwidth]{NDVI-rest}
		\label{fig:NDVI-C}
	\end{subfigure}
	\begin{subfigure}[b]{0.32\textwidth}
		\includegraphics[page=3,width=\textwidth]{NDVI-rest}
		\label{fig:NDVI-D}
	\end{subfigure}
	\begin{subfigure}[b]{0.33\textwidth}
		\includegraphics[page=4,width=\textwidth]{NDVI-rest}
		\label{fig:NDVI-E}
	\end{subfigure}
	\begin{subfigure}[b]{0.33\textwidth}
		\includegraphics[page=5,width=\textwidth]{NDVI-rest}
		\label{fig:NDVI-F}
	\end{subfigure}
\caption{The KDEs of the $448$ NDVI images, plus the FSSA plots for the grouping steps of the NDVI dataset.}
\label{fig:ndvi_group}
\end{center}
\end{figure}

\noindent Just like in \cite{haghbin2019}, we apply FSSA to the $448$ NDVI densities with a lag of $45$ and we obtain the exploratory plots of Figure \ref{fig:ndvi_group}.

\noindent Plot (A) of Figure \ref{fig:ndvi_group} shows a 3-D plot of the NDVI densities, (B) and (C) show that there are four components that don't correspond to noise, and plots (D), (E), and (F) show the existence of yearly periodicity in components 2 and 3 while component four captures a trend behavior in the data. We see that component four is a trend because after $2010$ (210 on x axis of plot (D)), we switch from weighting the fourth left singular function seen in plot (F), positively, to negatively such that instead of weighting heavily around 0.60, we weight more heavily around NDVI values of 0.4.

\subsection*{Approach of Hyndman and Ullah to FTS Forecasting}
The approach \cite{hyndman2007} leverages a weighted FPCA of a FTS to find basis elements that explain variation. They then project their time dependent data onto the basis in order to find scores for each principal component. Finally, they perform forecasting techniques such as ARIMA on the resulting principal component scores in order to predict future observations of the FTS. The following steps outline the process.

\begin{enumerate}
\item Smooth the FTS using a nonparametric smoothing method to estimate $f_{t}\left(x\right)$ for $x\in\left[x_{1},x_{p}\right]$ from sampling points $\{x_{i},y_{t}\left(x_{i}\right)\}$, $i=1,\dots,p$ where $f_{t}\left(x\right)$ is a function observed on day $t$ and $p$ is the number of sampling points
\item Decompose the fitted observations using a weighted FPCA to obtain the basis expansion of

\begin{equation*}
f_{t}\left(x\right)=\mu\left(x\right)+\sum_{k=1}^{K}\beta_{t,k}\phi_{k}\left(x\right)+e_{t}\left(x\right)
\end{equation*}

where $\mu\left(x\right)$ is the mean function, $\{\phi_{k}\left(x\right)\}_{k=1}^{K}$ is a set of orthonormal basis functions (notice we truncate at $K$ basis elements), $e_{t}\left(x\right) \sim \text{N}\left(0,\nu\left(x\right)\right)$, and $\nu\left(x\right)$ is the covariance function

\item Fit univariate time series models, such as ARIMA, to each of the coefficients $\{\beta_{t,k}\}_{k=1}^{K}$

\item Forecast the coefficients $\{\beta_{t,k}\}_{k=1}^{K}$ for $t=n+1,\dots,n+h$ using the fitted time series models where $n$ is the length of the FTS and $h$ is the forecast horizon

\item Multiply the forecasted coefficients by their respective $k^{\text{th}}$ basis elements to forecast $\{f_{n+1}\left(x\right),\dots,f_{n+h}\left(x\right)\}.$

\end{enumerate}

\end{document}